\documentclass[preprint,3p,times]{elsarticle}
\usepackage{graphicx}
\usepackage{ecrc}
\usepackage{lipsum}
\usepackage{amsmath} 
\usepackage{xcolor}
\usepackage{algorithm}
\usepackage{algpseudocode}
\usepackage{fancyvrb}

\usepackage{comment}

\usepackage{subcaption}
\usepackage{listings}
\usepackage{hyperref}
\usepackage{amssymb}
\AtBeginEnvironment{tcolorbox}{\small}
\usepackage[most]{tcolorbox}
\usepackage{float}
\usepackage{ulem}

\usepackage{tikz}
\usepackage{array}

\definecolor{darkblue}{rgb}{0.15,0.15,0.55}
\definecolor{lightgrey}{rgb}{0.75,0.75,0.75}

\firstpage{1}

\usepackage[figuresright]{rotating}

\begin{document}

\begin{frontmatter}

\title{Understanding Physical Properties of Unseen Deformable Objects by Leveraging Large Language Models and Robot Actions}

\author[1]{Changmin Park}
\ead{cmp9877@gmail.com}

\author[1]{Beomjoon Lee}
\ead{beomjoon.b.lee@gmail.com}

\author[1]{Haechan Jung}
\ead{chan993166@gmail.com}

\author[2]{Haejin Jung}
\ead{qpfh31@gmail.com}

\author[1]{and Changjoo Nam\corref{cor1}}
\ead{cjnam@sogang.ac.kr}

\affiliation[1]{organization={Dept. of Electronic Engineering, Sogang University},
            addressline={Baekbeom-ro, Mapo-gu}, 
            city={Seoul},
            postcode={04107}, 
            country={South Korea}}
\affiliation[2]{organization={Dept. of Mechanical Engineering, Korea Aerospace University},
            addressline={Hanggongdaehak-ro, Deogyang-gu}, 
            city={Goyang},
            postcode={10540}, 
            country={South Korea}}
            
\cortext[cor1]{Corresponding author}

\address{}

\begin{abstract}
In this paper, we consider the problem of understanding the physical properties of unseen objects through interactions between the objects and a robot. Handling unseen objects with special properties such as deformability is challenging for traditional task and motion planning approaches as they are often with the closed world assumption. Recent results in Large Language Models (LLMs) based task planning have shown the ability to reason about unseen objects. However, most studies assume rigid objects, overlooking their physical properties.
We propose an LLM-based method for probing the physical properties of unseen deformable objects for the purpose of task planning. For a given set of object properties (e.g., foldability, bendability), our method uses robot actions to determine the properties by interacting with the objects. Based on the properties examined by the LLM and robot actions, the LLM generates a task plan for a specific domain such as object packing. In the experiment, we show that the proposed method can identify properties of deformable objects, which are further used for a bin-packing task where the properties take crucial roles to succeed. 
\end{abstract}


\end{frontmatter}


\section{Introduction}~\label{sec:intro}

For robots operating in unseen and unstructured environments, the ability to understand their surroundings is crucial for their autonomous execution. One of representative approaches for achieving autonomy is task planning. Classical AI planning aims to create a sequence of transitions to achieve a predefined goal by abstracting the semantic information of objects and actions into states. They often use formalized languages such as STRIPS~\cite{fikes1971strips} or PDDL~\cite{aeronautiques1998pddl}. Many recent works use Large Language Models (LLMs) for task planning which have the advantages of commonsense knowledge and comprehensive power~\cite{liang2023code, ahn2022can, huang2022inner}. LLMs have shown remarkable performance in context understanding of physical relationships of objects~\cite{ma2024llms, fu2024scene}, which allow them to be used in robotic task and motion planning (TAMP)~\cite{wang2024llm}.


LLM-based task planning is particularly effective for robots operating in unstructured real-world environments such as domestic settings. However, existing task planning methods (e.g.,~\cite{huang2022inner, shirai2024vision, zhao2024large}) assume known objects. While there have been few recent works on LLM-based task planning with unseen objects~\cite{wu2023embodied, zhao2024large}, they do not consider deformability of objects. Since understanding the deformability requires a large amount of data to learn~\cite{wu2023learning,wang2023deri,deng2024learning} or an analytic model of object dynamics~\cite{guler2015estimating, petit2017using, haouchine2018vision}, existing methods restrict their coverages to a limited set of objects~\cite{zhou2024reactive, huang2023deformable} or learn how to manipulate deformable objects in an end-to-end fashion~\cite{matas2018sim} in the control level. To the best of our knowledge, there has been no existing work that can reason about the properties of unseen deformable objects to use them in long-horizon task planning.

\begin{figure}
\captionsetup{skip=0pt}
    \centering
    \includegraphics[width=0.47\textwidth]{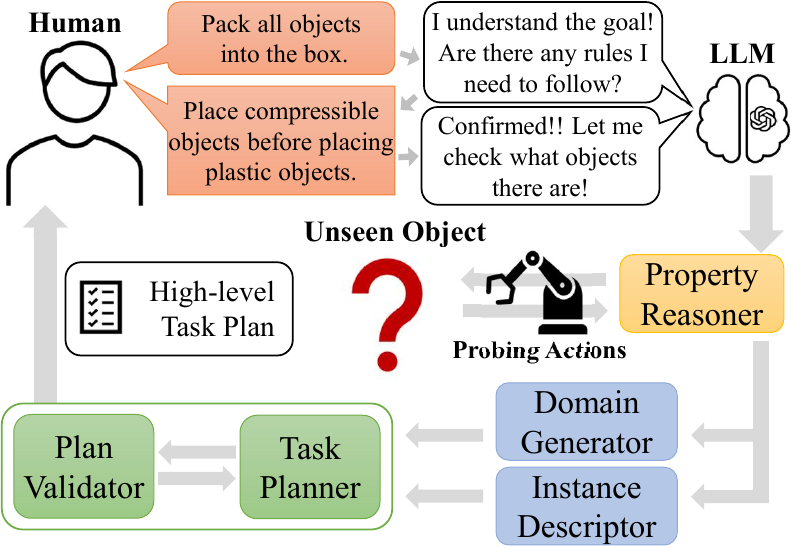}
    \caption{An overview of the proposed method for understanding physical properties of unseen objects by using the commonsense knowledge of an LLM and the interactability of robots.}
    \label{fig:overview}
\end{figure} 

In this paper, we propose an LLM-based method that can understand the physical properties of unseen objects (e.g., compressibility, bendability, plasticity) by observing interactions between an object and a robot. We show that the discovered properties can be used to generate task plans for a downstream task such as object bin-packing. 

As shown in Fig.~\ref{fig:overview}, our method consists of (i) the \textit{Property Reasoner} that detects objects and understands their properties using a visual language model (VLM)~\cite{liu2023grounding} along with an LLM (i.e., GPT-4o), (ii) the \textit{Domain Generator} which produces actions for task planning and predicates describing object properties, (iii) the \textit{Instance Descriptor} which generates a problem instance including the initial and goal states, and (iv) the \textit{Task Planner} that generates a task plan to perform a downstream task where errors in the plan are detected by the \textit{Plan Validator}. 

The main contributions include:
\begin{itemize}
    \item We propose a method that leverages the commonsense knowledge of LLMs to understand the physical properties of unseen deformable objects by observing a robot which interacts with objects.
    \item We present a protocol outlining how the robot interacts with objects, along with LLM prompts to help identify their properties.
    \item We develop a pipeline for fully automated task planning that adapts to the discovered properties of previously unseen objects.
\end{itemize}

\section{Related Work}~\label{sec:related}

Research on manipulating deformable objects has traditionally been studied at the motion level~\cite{wu2023learning, matas2018sim}, which needs prior knowledge of the physical properties of the objects. Based on the knowledge, they control robots to guide the objects toward the goal state. 
Learning-based methods~\cite{wang2019learning, yan2020self, lippi2020latent, deng2024learning} are proposed which have the ability to understand physical properties of deformable objects with a priori. For example, \cite{fu2024flingflow} proposes a method for unfolding clothes that learns the state classification and region segmentation of folded objects to guide the unfolding process. 
However, all the aforementioned methods are not easily applicable to unseen objects as they need prior knowledge or laborious data collection and training. While few methods (e.g., \cite{shridhar2022cliport, bartsch2024llm}) leverage foundation models to manipulate unseen deformable objects, they work in an end-to-end manner without understanding the physical properties of the objects, which can be further used for long-horizon task planning.

Research on automated generation of  knowledge base enables an autonomous agent to explore unseen environments. Generally, the agent expands a separated knowledge base through reasoning about its environment. For example, \cite{hanheide2017robot} proposes a knowledge hierarchy which allows assumption-based reasoning in uncertain environments. On the other hand, recent works leverage foundation models to recognize unseen objects from visual observations~\cite{song2023llm, ahn2024autort,shridhar2020alfred}. 
In~\cite{liu2024delta}, an LLM is used to generate domain knowledge for a given task from natural language instructions, where the knowledge is further used for task planning. 
Although these methods demonstrate the ability to generalize to unseen environments and objects, they may inaccurately predict the properties of objects because they rely on predefined object sets or built-in commonsense knowledge of LLMs.
Consequently, their ability to understand the physical properties of unseen deformable objects is limited due to their lack of capability to interact with and reason about their surroundings.

Recently, there has been a growing interest in identifying the physical properties of objects. 
Using fine-tuned LLMs trained on curated datasets, one line of research focus on estimating object properties by mapping target objects to known categories and predicting its properties based on similar examples.
For instance,~\cite{gao2024physically} demonstrates the limitations of VLMs in capturing physical characteristics and introduces a dataset specifically designed to address this gap. Similarly,~\cite{xie2024deligrasp} proposes a grasping strategy to estimate properties such as friction, mass, and spring constant. These approaches require additional training for high-level reasoning and often struggle to generalize across diverse object categories.
Another line of research focus on infering physical properties through direct robot-object interactions. For example,~\cite{zhao2023chat} leverages LLMs in conjunction with multi-modal sensors including sound, torque, and tactile data to reasoning the materials. Likewise,~\cite{lai2024vision} proposes a multi-modal reasoning method that integrates visual and haptic signals, using a robotic shaking action to identify liquid-containing objects. Despite their promise, these methods often depend on prior knowledge of specific physical properties and require additional training, which limits their adaptability to diverse and unstructured environments.

Our goal is to address the challenge of understanding the physical properties of unseen deformable objects without fine-tuning, enabling their use in long-horizon task planning. We aim to combine the commonsense reasoning of LLMs with physical interactions between a robot and objects.

\section{Problem Formulation}~\label{sec:PF}
Our goal is to understand physical properties of unseen objects to generate a grounded robot task plan for $\mathcal{T}(\mathcal{L} \mid \mathcal{C}) = (l_1, l_2, \cdots)$ from a language instruction \(\mathcal{L}\) (e.g., ``put all objects in the box'') under constraints \(\mathcal{C}\) (e.g.,``place a plastic object on top of a soft object''). This task plan transitions from the initial state $S_{\text{init}}$ to the goal state $S_{\text{goal}}$, where $l$ represents a high-level robot action. 
The planning structure consists of two main components: (i) a domain description $\textsc{Domain}$, which includes an object description $D$ containing predicates and action description $A$, and (ii) a description of a problem instance $\textsc{Problem}$ including $S_{\text{init}}$ and $S_{\text{goal}}$.  
Note that $A$ is the set of lifted actions (action schema) $A = \{\psi^g_1, \psi^g_2, \cdots\}$, which are generated from action primitives $\Psi = \{\psi_1, \psi_2, \cdots \}$ (e.g., pick(object, spot)). The task plan \(\mathcal{T}(\mathcal{L} \mid \mathcal{C})\) can be expressed as:
\[\mathcal{T}(\mathcal{L} \mid \mathcal{C}) = \textsc{Planner}(\textsc{Domain}, \textsc{Problem})\]
where $\textsc{Domain} = (D, A)$ and $\textsc{Problem} = (S_{\text{init}}, S_{\text{goal}})$. For simplicity, we use $\mathcal{T}$ interchangeably with $\mathcal{T}(\mathcal{L} \mid \mathcal{C})$. 

To generate $\mathcal{T}$ to manipulate the objects, we want to discover the physical properties of them, which begins with detecting a set of $N$ objects. A set of tuple $\textbf{o} = (o, \mathcal{X})$ formulates detected objects $\mathcal{O}=\{\textbf{o}_1,\textbf{o}_2,\cdots,\textbf{o}_N\}$  where $\mathcal{X}$ includes physical properties of the detected object $o$. 
Other visual information of $\mathcal{O}$ (e.g., color, dimension, and shape) can be obtained from $M$ images \(\mathcal{I} = \{i_1, i_2, \cdots, i_M\}\). For brevity, we assume that those visual information of objects is stored in $o$ without dedicated symbols.
Since detection relies on visual information, we assume $\textbf{o} = (o, \mathcal{X} = \emptyset)$ at this stage, which means that the physical properties of the objects cannot be inferred from the images only.

We want to construct a knowledge base $\mathcal{K} = \{ (o_1, \mathcal{X}_1), (o_2, \mathcal{X}_2), \cdots \}$ where $\mathcal{X} \neq \emptyset$ for all objects. Initially, the knowledge base could be empty or partially filled with existing knowledge. An object $\textbf{o}$ is considered unseen if $\textbf{o} \notin \mathcal{K}$. A probing procedure $f_p$ can grow the knowledge $\mathcal{K} \leftarrow \mathcal{K} \cup \{\textbf{o}^\prime = (o, \mathcal{X} \neq \emptyset)\} $ where $\textbf{o}^\prime = f_p(\textbf{o})$ indicates that the physical properties of the detected object $o$ can be obtained by $f_p$. The finite set of robot probing actions $\mathcal{A}_p = \{a_{p_1}, a_{p_2}, \cdots \}$\footnote{Notice that the probing actions are irrelevant to $A$, which is the set of actions used in task planning.} construct $f_p$. Once all objects in $\mathcal{O}$ become known, the probing procedure terminates. 


To determine $\mathcal{T}(\mathcal{L} \mid \mathcal{C})$, we aim to develop a unified framework for automated task planning with the capabilities for (i) discovering properties of unseen objects, (ii) generating a planning instance and a domain knowledge for the instance, and (iii) producing a valid task plan incorporating the discovered properties of the unseen objects.

\section{Method}~\label{sec:method}
As shown in Fig.~\ref{fig:framework}, we develop a framework which consists of the property reasoner, domain generator, instance descriptor, task planner, and plan validator. The first three components generate a planning instance and a domain knowledge whereas the last two generate and validate task plans. 

\subsection{Property Reasoner}~\label{sec:reasoner}
\vspace{-20pt}
\subsubsection{Object Detection and Naming}~\label{sec:OD}
\begin{figure*}[t]
    \centering
    \includegraphics[width=\linewidth]{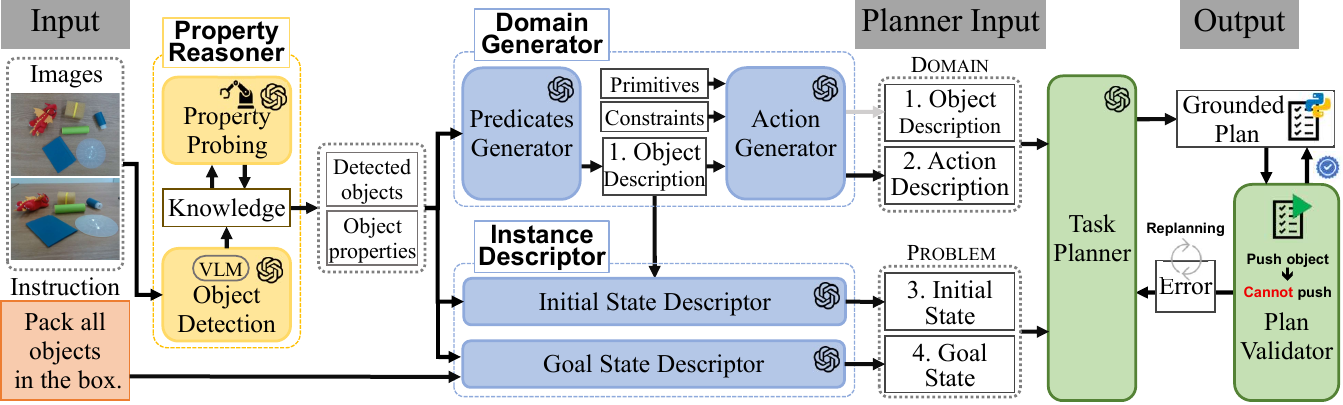}
    \caption{The overall procedure of our method for the bin-packing domain involves autonomously investigating object properties to expand domain knowledge using a robot. The property reasoner gathers visual information about objects (e.g., color, dimensions, shape). If the knowledge base lacks the physical properties of the objects, the reasoner uses images of the objects interacting with the robot to probe these properties. Using the object information and a language instruction, the domain generator and instance descriptor create structured data for task planning. The generated task plan is then validated by the plan validator to check for syntax and semantic errors.
    }
    \label{fig:framework}
    \vspace{-0.4cm}
\end{figure*} 
The property reasoner detects \(\mathcal{O}\) from $\mathcal{I} = \{ i_1, \cdots, i_M \}$ and probes their properties where the images in $\mathcal{I}$ capture $\mathcal{O}$ from $M$ different perspectives to accurately detect the dimension and shape.\footnote{We set $M=2$  where more images could provide more accurate information at a cost.} 
To let the LLM clearly distinguish objects, we find bounding boxes of objects using a VLM~\cite{liu2023grounding}. In addition, we draw a topological graph where each vertex is the center of the bounding box of an object. Three nearest vertices are connected to each vertex. 
After all objects are found, GPT-4o~\cite{achiam2023gpt} names all objects in the Chain of Thought~\cite{wei2022chain} manner. Since an unseen object is unlikely to have a proper name, its label is generated by concatenating descriptive information about the object, such as { \texttt{green\_3D\_cylinder}}. We note that the prompts for the LLM (\ref{app:od}) consist of the processed images from the VLM and the definitions of shape, dimension, and property, $\mathcal{D}$. The definitions are obtained from the Cambridge Dictionary~\cite{cambridge_dictionary} and provided in~\ref{app:Def}. 

\subsubsection{Property Probing}~\label{sec:property_probing}
The physical properties of detected objects are reasoned by the LLM where the objects interact with a robot physically. If the number of properties is excessive, the LLM is likely to have hallucinations. Thus, we limit the set of object properties depending on the task domain. As a running example, we use the bin-packing domain where a set of objects are packed into a bin. In this domain, we focus on deformability of objects to pack objects while preventing irreversible deformation of objects. To do so, we introduce a set of five properties which are rigidity, bendability, foldability, compressibility, and plasticity. An object can be classified as either rigid or non-rigid based on whether it maintains its shape when subjected to external forces. Non-rigid objects can exhibit deformation properties that depend specifically on their dimensions. 1D objects such as a string or a needle can be bendable. 2D objects like a sheet of paper or a dish plate can be either foldable. 3D objects such as sponges can be compressed. If a deformed (i.e., bent, folded, and pushed) object cannot be recovered to its original state, it is plastic deformable (plastic in short) regardless of its dimension.  

The properties impose constraints or objectives for safe and space-efficient bin packing: (i) Bendable (1D) objects can be bent to fit the size of the bin. (ii) Foldable (2D) objects can be folded to utilize the bin space as much as possible. (iii) Compressible (3D) objects can be compressed to secure more space for packing. Also, they can be used to protect objects that can have plastic deformation. (iv) Plastic objects can be easily damaged. Protecting them using compressible objects can be beneficial for packing tasks. Note that rigid objects can be placed in a bin without constraints.

\begin{figure}[t!]
    \centering
    \begin{subfigure}[b]{0.50\textwidth}
        \captionsetup{skip=0pt}
	    \includegraphics[width=\textwidth]{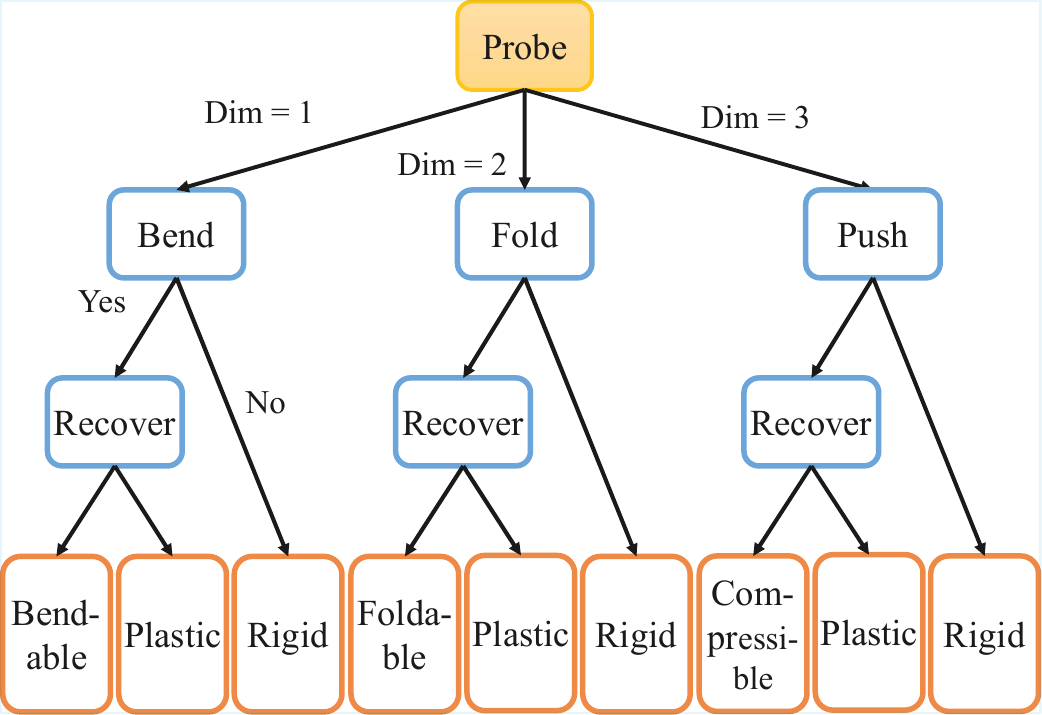}
        \caption{A decision tree to determine the property of an object}
        \label{fig:tree}
    \end{subfigure}
    \quad
    \begin{subfigure}[b]{0.45\textwidth}
        \captionsetup{skip=0pt}
	    \includegraphics[width=\textwidth]{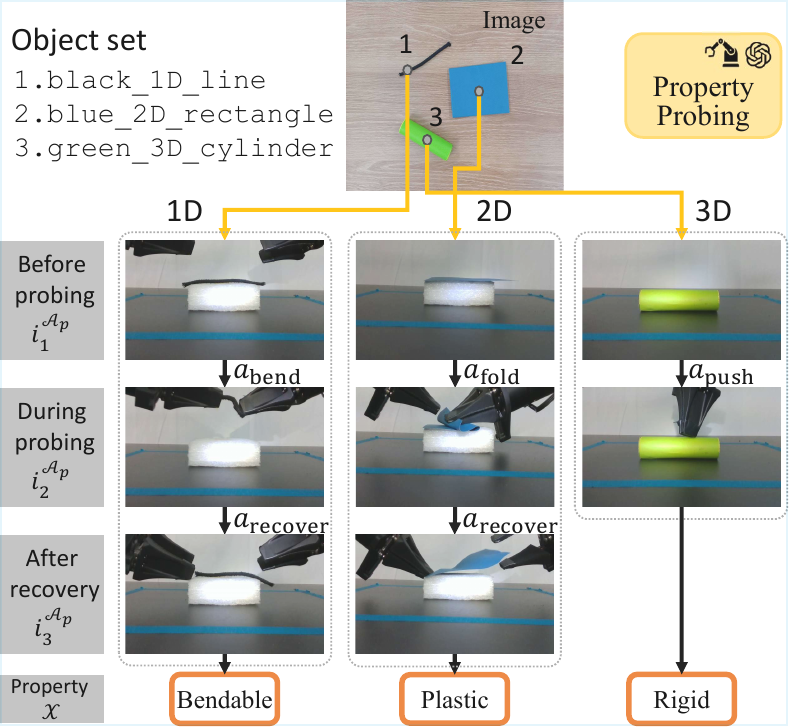}
        \caption{An example use of the tree}
        \label{fig:action}
    \end{subfigure}
    \caption{The decision tree approach for determining object properties. (a) By applying a series of actions, one of the properties at the leaf nodes is determined. (b) Depending on the dimension of objects, a dual-arm robot performs appropriate probing actions. For example, the property of { \texttt{blue\_2D\_rectangle}} is \textit{plastic} as the object is foldable but not recovered to its original state even after an unfolding action.}
  \label{fig:probing_property}
\end{figure}

To probe these properties, we define a set of probing actions $\mathcal{A}_p = \{a_{\text{bend}}, a_{\text{fold}}, a_{\text{push}}, a_{\text{recover}}\}$ which are performed by a dual-arm robot.
As described in Fig.~\ref{fig:action}, the robot applies a bending action \( a_{\text{bend}} \) to 1D objects by grasping both ends of the object and attempting to bend it until one end meets the other. 
Similarly, the robot applies a folding action $a_{\text{fold}}$ to 2D objects by grasping one endpoint on the right and pressing the opposite endpoint, attempting to fold the object until the endpoints meet. 
The robot applies a pushing action $a_{\text{push}}$ to 3D objects by pushing the center of the object from the top. After applying each of these deforming actions (e.g., $a_{\text{bend}}$, $a_{\text{fold}}$, and $a_{\text{push}}$), the robot applies a recovery action $a_{\text{recover}}$ for those objects that are not rigid. The resulting states of the objects can determine whether they are plastic or not. Three images are captured during property probing: $i_{1}^{\mathcal{A}_p}$ before the deforming action, $i_{2}^{\mathcal{A}_p}$ during the deforming action, and $i_{3}^{\mathcal{A}_p}$ after a recovery action. 

We develop two approaches to utilize images for reasoning with the LLM. In the first approach, the LLM is directly queried about the properties using three images, with each image labeled to indicate the action that produced it. The second approach employs a decision tree (Fig.~\ref{fig:tree}) based on property definitions provided by $\mathcal{D}$. An example of this process is shown in Fig.~\ref{fig:action}. The label {\texttt{blue\_2D\_rectangle}} in $\mathcal{O}$ is assigned to a blue paper sheet in the image, whose properties are initially unknown. In $i_{1}^{\mathcal{A}_p}$, it appears to be flat. Since it is recognized as a 2D object, the robot performs $a_{\text{fold}}$. The LLM then recognizes the object as folded in $i_{2}^{\mathcal{A}_p}$. Finally, the robot applies $a_{\text{recover}}$ to unfold the object. Since the object does not return to its original flat shape, its property is determined to be plastic.

Our overall procedure for the property reasoning method is described in Alg.~\ref{alg:property_reasoner} and prompts are provided in~\ref{app:pp}. In line~\ref{alg_probing:detection}, the VLM draws bounding boxes of objects in $\mathcal{I}$. The processed images are sent to the LLM to label the objects to obtain $\mathcal{O}$. If an object in $\mathcal{O}$ is already in $\mathcal{K}$, its properties are retrieved from $\mathcal{K}$ (line~\ref{alg_probing:seen_obj}). Otherwise, the probing process $f_p$ is executed (line~\ref{alg_probing:unseen_obj}). In the latter case, an unseen object turns into a known object so $\mathcal{K}$ is augmented with that object. Finally, $\mathcal{O}$ is returned with all objects known.

\setlength{\textfloatsep}{4pt}
\begin{algorithm}[t]
    \caption{Property Reasoner}~\label{alg:property_reasoner}
    \begin{algorithmic}[1]
\State \textbf{Input}: Images \(\mathcal{I} = \{i_1, i_2, \cdots, i_M\}\), Knowledge Base \(\mathcal{K}\), definition $\mathcal{D}$
    \State \textbf{Output}: a set of current objects $\mathcal{O}$ 
    \State $\mathcal{O} \gets$ \textsc{LLM}(\textsc{VLM}$(\mathcal{I}) \mid \mathcal{D})$  \label{alg_probing:detection}
    \ForAll{$\textbf{o} = (o, \emptyset) \in \mathcal{O}$} 
        \If{$o$ is found in $\mathcal{K}$}\label{alg_probing:seen_obj}
            \State Retrieve $\mathcal{X}$ of $o$ from $\mathcal{K}$
            \State $\textbf{o}^\prime \gets (o, \mathcal{X})$
        \Else\label{alg_probing:unseen_obj}
            \State $\textbf{o}^\prime \gets f_p(\textbf{o})$ \label{alg_probing:probing} \Comment{Probing the property of $\textbf{o}$}
            \State $\mathcal{K} \gets \mathcal{K} \cup \{\textbf{o}^\prime\}$ \Comment{Update the knowledge base.}
        \EndIf
        \State Replace $\textbf{o} \in \mathcal{O}$ by $\textbf{o}^\prime$ 
    \EndFor{}
    \Return $\mathcal{O}$
\end{algorithmic}
\end{algorithm}

\subsection{Task Planning}
\subsubsection{Domain Generator}~\label{sec:DG}
Task planning requires a domain description (\textsc{Domain}) and a problem instance description (\textsc{Problem}). The domain description specifies which actions can be applied to which objects, while the instance description includes the initial state and the goal state of a particular planning instance. In our implementation, the LLM is asked to generate two descriptions in Python as task planning is done by executing the Python script including the descriptions. 

A domain description $\textsc{Domain} = (D, A)$ consists of the object description $D$ and the action description $A$ which is generated to comply with the constraint set $\mathcal{C}$. We assume that $\mathcal{C}$ is given from a domain expert or commonsense knowledge, where our bin-packing task has constraints to protect plastic objects from deformation and use the bin space frugally: (i) Before placing a plastic object, a compressible object should be in the box before. Non-plastic objects can be placed without any constraints. (ii) Push all compressible objects after placing them in the box. (iii) If there is a bendable object, bend it before placing it in the box. (iv) If there is a foldable object, fold it before placing it in the box. (v) Do not bend, fold, or push a plastic object.

The domain generator receives the knowledge obtained by the property reasoner. It uses the predicate generator and the action generator (which belong to the domain generator in Fig.~\ref{fig:framework}) to generate $D$ and $A$, respectively. From $\mathcal{O}$, the predicate generator produces $D$ which contains a set of predicates describing objects and their states. Each $\mathcal{X}$ of $\textbf{o}^\prime \in \mathcal{O}$ is converted into predicates such as {\small \texttt{is\_rigid}} or {\texttt{is\_plastic}} by prompting the LLM (the prompts are provided in~\ref{app:pg}). In addition to the predicates describing object properties, $D$ includes domain-dependent predicates such as {\texttt{in\_bin}}. This process can be expressed as: $D=\textsc{LLM}(\mathcal{O})$.

The action generator produces the action description $A$ from $D$, given action primitives $\Psi=\{\psi_1, \psi_2, \cdots\}$ and $\mathcal{C}$ (i.e., $A=\textsc{LLM}(D \mid \Psi, \mathcal{C})$). The action primitives are simply abstract-level manipulation actions. In our bin-packing task, $\Psi$ consists of primitive actions \textit{pick}, \textit{place}, \textit{bend}, \textit{fold}, and \textit{push}. The LLM converts the action primitives into lifted actions (action schema) $\psi_i^g$ which are described by their parameters, the required states to perform the actions (i.e., preconditions), and the consequences of the actions (i.e., effects). The preconditions and effects are generated to comply with all constraints given by $\mathcal{C}$. The prompts for generating $A$ are given in~\ref{app:ag}.

\subsubsection{Instance Descriptor}~\label{sec:ID}
An instance description $\textsc{Problem}=(S_{\text{init}}, S_{\text{goal}})$ consists of the initial state and the goal state by the initial state descriptor and the goal state descriptor, respectively (which belong to the instance generator in Fig.~\ref{fig:framework}). From $\mathcal{O}$ and $D$, the initial state descriptor generates $S_{\text{init}}$ which represents the current states of $\textbf{o}^\prime$ (i.e., $S_{\text{init}}=\textsc{LLM}(\mathcal{O}, D)$). 
The goal state descriptor generates $S_{\text{goal}}$ from $\mathcal{O}$ and $\mathcal{L}$ (i.e.,  $S_{\text{goal}}=\textsc{LLM}(\mathcal{O}, \mathcal{L})$). While $S_{\text{init}}$ is in Python, $S_{\text{goal}}$ is in a table to concisely summarize the resulting state after executing the task plan. The prompts to the LLM for the instance generation are provided in~\ref{app:is} and \ref{app:gs} where example results are also given in the prompts.

\subsubsection{Task Planner and Plan Validator}~\label{sec:TP_PV}
By the domain generator and instance generator, all necessary components for task planning are prepared. The task planner generates a grounded task plan, while the plan validator checks its validity. If the plan is found to be invalid, it is fed back to the task planner for replanning. 

The LLM as the task planner produces a plan $\mathcal{T}=(l_1, l_2, \cdots, l_N)$ in Python, using $\textsc{Domain} = (D, A)$ and $\textsc{Problem}=(S_\text{init}, S_\text{goal})$. The Python functions invoked in the task plan are the grounded actions of the lifted actions defined in $A$. Together with $S_\text{init}$, an executable Python script is completed. The prompts for task planning are shown in~\ref{app:tp} where an example Python script is shown in~\ref{app:TP_Result}.

The plan validator (also the LLM) examines the task plan by executing the Python script and looking at the execution result denoted by {\texttt{Exe}}. Errors in {\texttt{Exe}} indicate incorrectly generated task plans which are either a syntax error (i.e., grammar error) or a semantic error (i.e., constraint violation). If no syntax error is detected, the validator checks if the result contains an error tag\footnote{In our implementation, the tag starts with {\texttt{Cannot}} followed by the name of the inexecutable action as shown in~\ref{app:PV_Result}. The Python function corresponding to the action is defined to prompt the tag if the precondition is not satisfied.}. Even though a semantic error is detected, it is difficult to identify the source of the error which can be the action generation or task planning. An action might be generated not to comply with the constraints given by $\mathcal{C}$. Otherwise, the order of the actions in the task plan might be incorrect. 

Therefore, if a semantic error is detected, we regenerate $A$ by providing {\texttt{Exe}} and then replan $\mathcal{T}$. This process can be expressed as $A^\prime=\textsc{LLM}(D, \texttt{Exe} \mid \Psi, \mathcal{C})$ and $\mathcal{T}^\prime=\textsc{LLM}(D, A, S_{\text{init}}, S_{\text{goal}}, \texttt{Exe})$. Our method repeats this validation process iteratively until $\mathcal{T}$ without any error is obtained (the second example in~\ref{app:PV_Result}).

\section{Experiments}~\label{sec:experiment}
\vspace{-10pt}

\begin{figure*}[t]
\captionsetup{skip=0pt}
    \centering
    \includegraphics[width=0.98\linewidth]{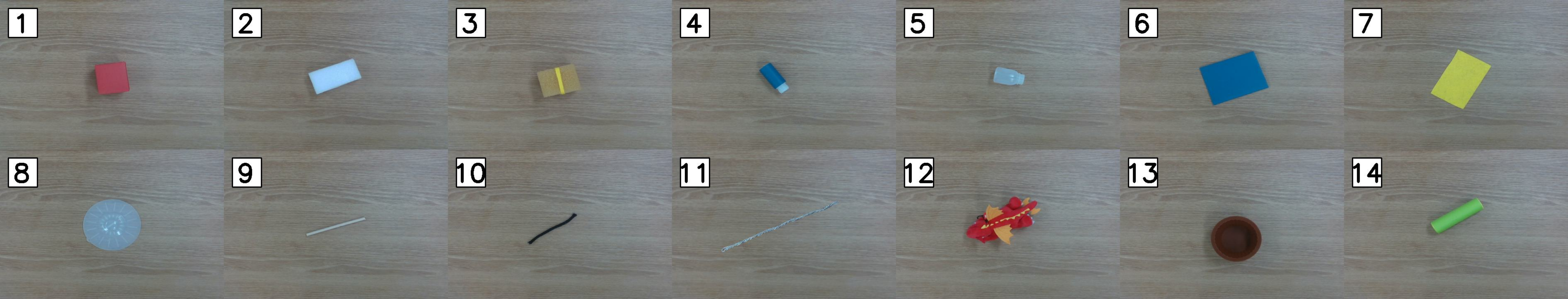}
    \caption{The set of 14 objects used in the experiments. The dimensions and physical properties of them are listed in the second and third columns of Table~\ref{tab:name_result}.}
    \label{fig:ex_objects}
\end{figure*}

We design a set of experiments to show the performance of the proposed method where metrics are (i) the accuracy of the property reasoner in probing the physical properties of objects and (ii) the success rate of the task planner given the physical properties of objects. 
The set of 14 objects used in the experiments is shown in Fig.~\ref{fig:ex_objects} including rigid and non-rigid objects with different dimensions. Our object set consists of objects that are difficult to recognize in their physical properties if visual information is solely given. For example, predicting the properties of Objects 1 and 2 is difficult due to a lack of visual features. Using selected objects, we generate 38 bin-packing random instances where each instance is with 3 to 7 objects. Since the LLM could produce different answers even for the identical prompt, we repeat testing the same instance ten times to obtain the statistics. 

\subsection{Settings}
We use two Agile-X 6-DOF manipulators for the experiments to perform property probing and task planning execution. The motions of the robots for property probing and task planning are generated by human teaching, which is called Programming by Demonstration (PbD). The LLM used in the experiments is GPT-4o where the parameters are set to the default values except for \textit{temperature} set to $0.2$ and \textit{top\_p} set to $0.7$ for text-only input. These values are set to $0.2$ and $0.1$ for prompting visual information.


\subsection{Property Reasoning}~\label{sec:property_reasoning}

The performance of property reasoning heavily depends on the performance of object detection and naming. Thus, we first measure the success rates of object detection and naming. Then we measure the accuracy of property probing for those objects that are successfully detected and named. 

\begin{figure}[t]
\captionsetup{skip=0pt}
    \centering
    \includegraphics[width=0.6\textwidth]{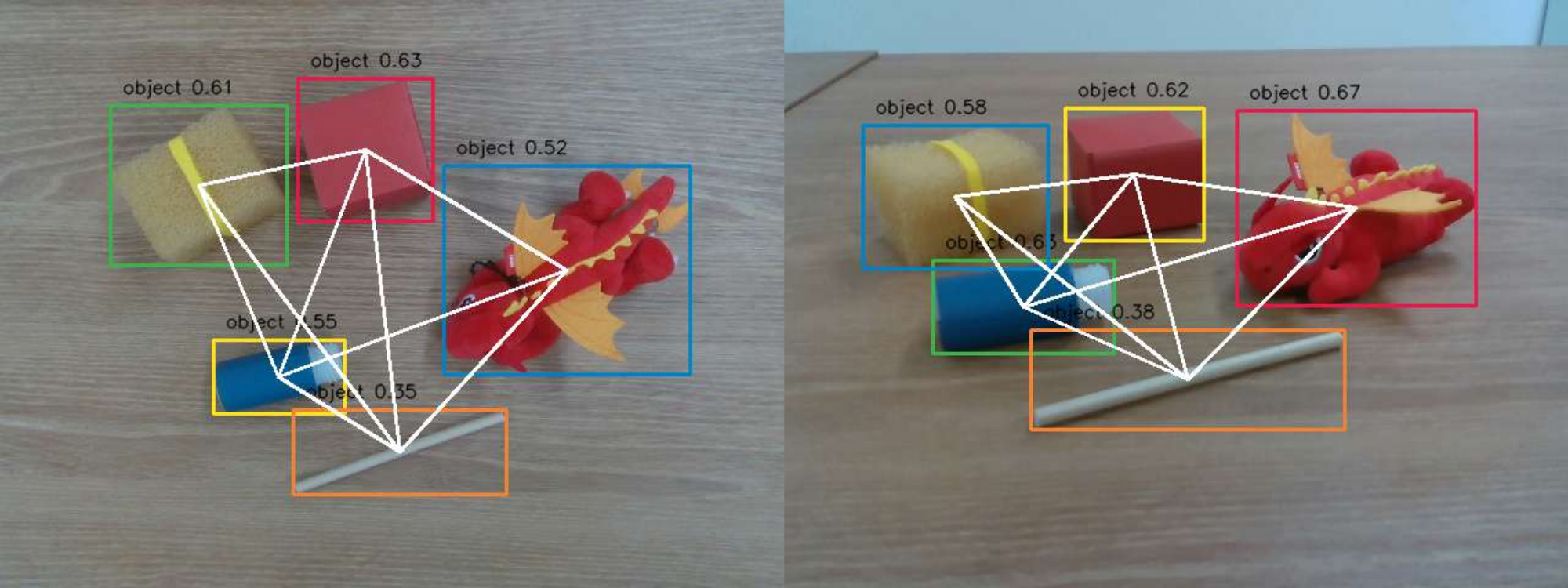}
    \caption{An example of a set of images processed by a VLM and a connection graph. Since the objects are unknown yet, their labels are determined arbitrarily. The graph helps the LLM understand the spatial relationship between objects which reduces the hallucination.}    
    \label{fig:scene_images}
\end{figure}

\begin{figure}[t]\label{fig:table_all}
    \captionsetup{skip=0pt}
    \centering
    \begin{subfigure}{0.425\textwidth}
        \captionsetup{skip=0pt}
        \includegraphics[width=\textwidth]{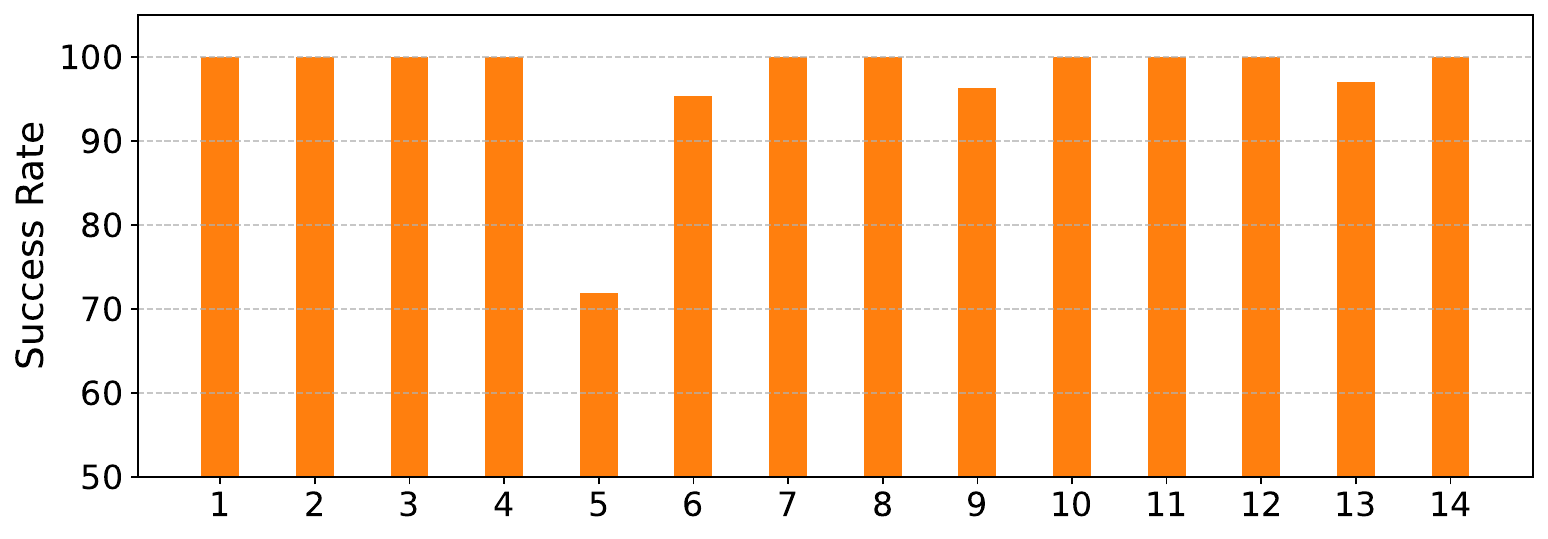}
        \caption{Object naming}
        \label{fig:naming}
    \end{subfigure}
    \begin{subfigure}{0.245\textwidth}
        \captionsetup{skip=0pt}
        \includegraphics[width=\textwidth]{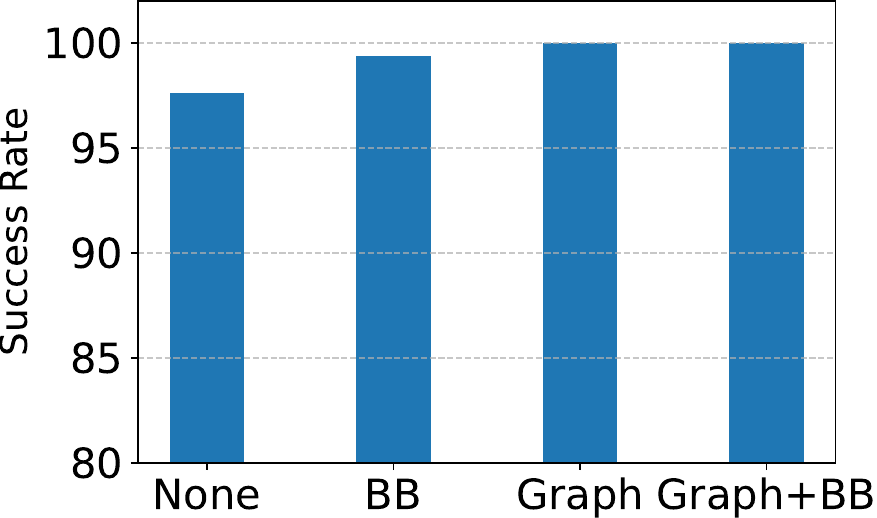}
        \caption{Object detection}
        \label{fig:detection}
    \end{subfigure}%
    \begin{subfigure}{0.26\textwidth}
        \captionsetup{skip=0pt}
        \includegraphics[width=\textwidth]{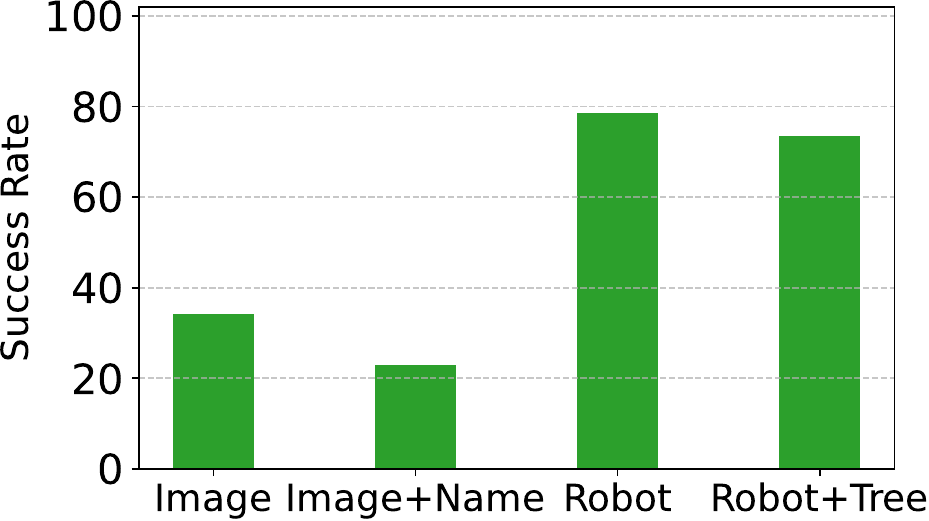}
        \caption{Property probing}
        \label{fig:probing}
    \end{subfigure}
    \caption{The success rates of the property reasoner and its components. (a) The high success rate of object naming shows that the LLM can recognize the shape and dimension of objects consistently. (b) The effectiveness of the additional visual information (the bounding boxes and graphs) is shown. (c) The physical interactions with objects greatly help in understanding the properties of objects.} \label{fig:pp}
\end{figure}

For each bin-packing instance among 38, our method uses two images from different perspectives that are top and side views as shown in Fig.~\ref{fig:scene_images}. The two images contain the same objects. The object detection is successful if all objects in the images are detected without any false positives. In other words, a successful object detection must not have a missing object and a hallucination. To test the effects of the bounding boxes and the graph, we compare four methods for object detection which are \textsf{None} (raw images), \textsf{BB} (bounding box only), \textsf{Graph} (graph only), and \textsf{BB+Graph} (both). 
Each instance is tested ten times with an identical prompt so a total of 380 trials. Fig.~\ref{fig:scene_images} shows the bounding boxes and graph. As summarized in Table~\ref{tab:od_result} and Fig.~\ref{fig:detection}, \textsf{Graph} and \textsf{BB+Graph} achieve 100.00\% of the success rate. While \textsf{None} and \textsf{Graph} achieve high success rates, providing additional visual information helps improve the detection performance of the LLM. Specifically, providing spatial information (i.e., using the graph) about the objects plays a crucial role in detecting missing objects and preventing hallucinations. 

We measure the success rate of object naming where objects are detected by \textsf{BB+Graph}. A successful naming should provide both the correct dimension and the shape of the object in the name. The object color is not considered as color is not used in the bin-packing task. The 38 instances have 184 objects in total including duplicates (i.e., the same object is counted twice if it is in two instances). By repeating ten trials for each instance, we have a total of 1840 resulting names to measure the success rate. As shown in Table~\ref{tab:name_result} and Fig.~\ref{fig:naming}, object dimensions and shapes are correctly identified in most cases. While most objects achieve success rates close to 100\%, Object 5 is occasionally misclassified as a cuboid (28.12\%) despite being cylindrical. We believe this is due to the ambiguity of the shape as it is not a perfect cylinder, as it features partial angulations. Its translucency and lack of texture add further complexity to the shape recognition process of the LLM.

Finally, we measure the performance of the property probing given the satisfactory performance of the object detection and naming. 
We have ten trials for each of 14 objects so 140 test cases. 
The deforming actions (e.g., $a_\text{bend}$, $a_\text{fold}$, and $a_\text{push}$) are individually taught using the PbD mathod. Each action is followed by a corresponding recovery action, executed sequentially within a single cycle.\footnote{Please refer to the~\ref{app:property_probing} and supplemental video for further details.}
The ground truth properties of the objects are given in Table~\ref{tab:pp_result}.
We compare four different methods. \textsf{Image} and \textsf{Image+Name} do not use the robot but use the LLM only for reasoning about the properties. 
While \textsf{Image} uses the image only, \textsf{Image+Name} uses the image and the object label found by a pretrained Vision Transformer~\cite{dosovitskiy2020image}. 
We also test methods including physical interactions of the objects with the robot. \textsf{Robot} uses the three images of the object which show the state of the object before, during, and after interactions with the robot. \textsf{Robot+Tree} uses the decision tree explained in Sec.~\ref{sec:property_probing}. We note that the image used in \textsf{Image} and \textsf{Image+Name} is the same as the before probing image used in \textsf{Robot+Tree} and \textsf{Robot+Tree}.

The result shown in Table~\ref{tab:pp_result} and Fig.~\ref{fig:probing} demonstrates that our interaction-based methods achieve accuracy up to 78.57\% across all objects. In contrast, the non-interaction methods, \textsf{Image} (34.29\%) and \textsf{Image+Name} (22.86\%), show a significant limitation in predicting physical properties as they do not involve physical interactions with the objects. The lower success rate of \textsf{Image+Name} indicates that providing object names to the LLM could bias the reasoning of it. 
The \textsf{Robot} (78.57\%) method outperforms \textsf{Robot+Tree} (73.57\%) slightly. This result indicates that using the decision tree could limit the flexibility of the reasoning process of the LLM. 
While the experimental results show that robot interactions significantly help the LLM reason about the object properties, imposing constraints (i.e., object names, decision trees) appears to restrict the reasoning capabilities of LLMs, leading to reduced performance.

\begin{table}[t]
\setlength\tabcolsep{4pt} 
\captionsetup{skip=0pt}
\caption{The success rates of the property reasoner and its components, with numerical results corresponding to Fig.~\ref{fig:pp}}
  \begin{subtable}{0.5\textwidth}
  \captionsetup{skip=0pt}
    \centering
    \caption{Object naming}
    \resizebox{0.99\columnwidth}{!}{
    \begin{tabular}{|c|c|c|c||c|c|c|c|}
    \hline
    Object & Dim & Shape & Total & Object & Dim & Shape & Total\\
    \hline
    1 & 100.00 & 100.00 & 100.00 & 8 & 100.00 & 100.00 & 100.00\\
    \hline
    2 & 100.00 & 100.00 & 100.00 & 9 & 96.36 & 100.00 & 96.36\\
    \hline
    3 & 100.00 & 100.00 & 100.00 & 10 & 100.00 & 100.00 & 100.00\\
    \hline
    4 & 100.00 & 100.00 & 100.00 & 11 & 100.00 & 100.00 & 100.00\\
    \hline
    5 & 100.00 & 71.88 & 71.88 & 12 & 100.00 & 100.00 & 100.00\\
    \hline
    6 & 95.38 & 100.00 & 95.38 & 13 & 100.00 & 97.06 & 97.06\\
    \hline
    7 & 100.00 & 100.00 & 100.00 & 14 & 100.00 & 100.00 & 100.00 \\
    \hline
    \end{tabular}\label{tab:name_result}}
  \end{subtable}
  \begin{subtable}{0.5\textwidth}
  \captionsetup{skip=0pt}
    \centering
    \caption{Object detection}
    \resizebox{0.99\columnwidth}{!}{
  \begin{tabular}{|c|c|c|c|}
    \hline
      & Success Rate (\%) & Missing objects (\%) & Hallucinations (\%) \\
    \hline
    \textsf{None} & 97.63 & 1.84 & 0.53 \\
    \hline
    \textsf{BB} & 99.37 & 0.00 & 0.53\\
    \hline
    \textsf{Graph} & 100.00 & 0.00 & 0.00 \\
    \hline
    \textsf{BB+Graph} & 100.00 & 0.00 & 0.00 \\
    \hline
    \end{tabular}\label{tab:od_result}}
  \end{subtable}
  
  \begin{subtable}{1\textwidth}
  \captionsetup{skip=0pt}
    \centering
    \caption{Property probing (R: rigid, B: bendable, F: foldable, C: compressive, P: plastic)}
    \resizebox{1\columnwidth}{!}{
    \begin{tabular}{|c|c|c|c|c|c|c||c|c|c|c|c|c|c|}
    \hline
    \multicolumn{3}{|c|}{Index} & \multicolumn{2}{c|}{No interaction} & \multicolumn{2}{c||}{Interaction} & \multicolumn{3}{c|}{Index} & \multicolumn{2}{c|}{No interaction} & \multicolumn{2}{c|}{Interaction} \\
    \hline
    Object & Dim & Properties & \textsf{Image} & \textsf{Image+Name} & \textsf{Robot} & \textsf{Robot+Tree} & Object & Dim & Properties & \textsf{Image} & \textsf{Image+Name} & \textsf{Robot} & \textsf{Robot+Tree} \\
    \hline
    1 & 3 & P & 0.00 & 0.00 & 100.00 & 100.00 & 8 & 2 & F & 0.00 & 0.00 & 100.00 & 100.00 \\
    \hline
    2 & 3 & R & 0.00 & 0.00 & 0.00 & 0.00 & 9 & 1 & R & 60.00 & 100.00 & 0.00 & 0.00 \\
    \hline
    3 & 3 & C & 80.00 & 20.0 & 100.00 & 100.00 & 10 & 1 &  B & 0.00 & 0.00 & 100.00 & 100.00 \\
    \hline
    4 & 3 & R & 100.00 & 100.00 & 100.00 & 100.00 & 11 & 1 &  P & 0.00 & 0.00 & 100.00 & 40.00 \\
    \hline
    5 & 3 & C & 100.00 & 0.00 & 100.00 & 100.00 & 12 & 3 &  C & 30.00 & 0.00 & 100.00 & 90.00 \\
    \hline
    6 & 2 & P & 00.00 & 0.00 & 0.00 & 40.00 & 13 & 3 &  C & 0.00 & 0.00 & 100.00 & 60.00 \\
    \hline
    7 & 2 & F & 90.00 & 0.00 & 100.00 & 100.00 & 14 & 3 &  R & 20.00 & 100.00 & 100.00 & 100.00 \\
    \hline
    \end{tabular}\label{tab:pp_result}}
  \end{subtable}
\end{table}

\begin{figure}[t!]
    \centering
    \begin{subfigure}[b]{0.53\textwidth}
        \captionsetup{skip=0pt}
	    \includegraphics[width=\textwidth]{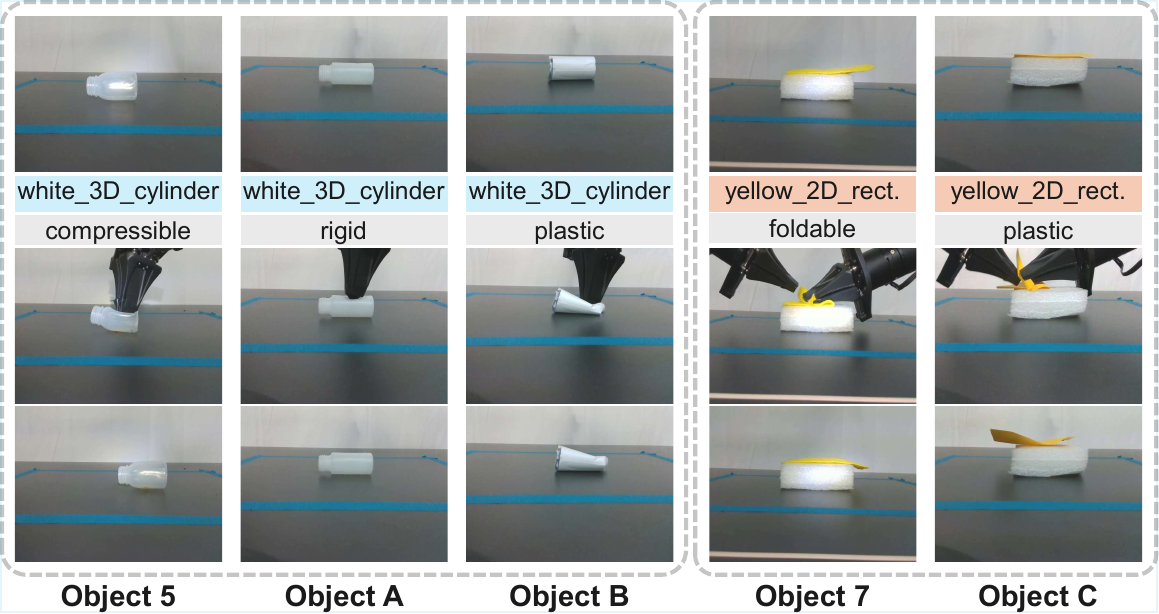}
        \caption{Property probing for objects with similar appearances}
        \label{fig:additional_property_probing_figs}
    \end{subfigure}
    \begin{subfigure}[b]{0.45\textwidth}
        \captionsetup{skip=0pt}
	    \includegraphics[width=\textwidth]{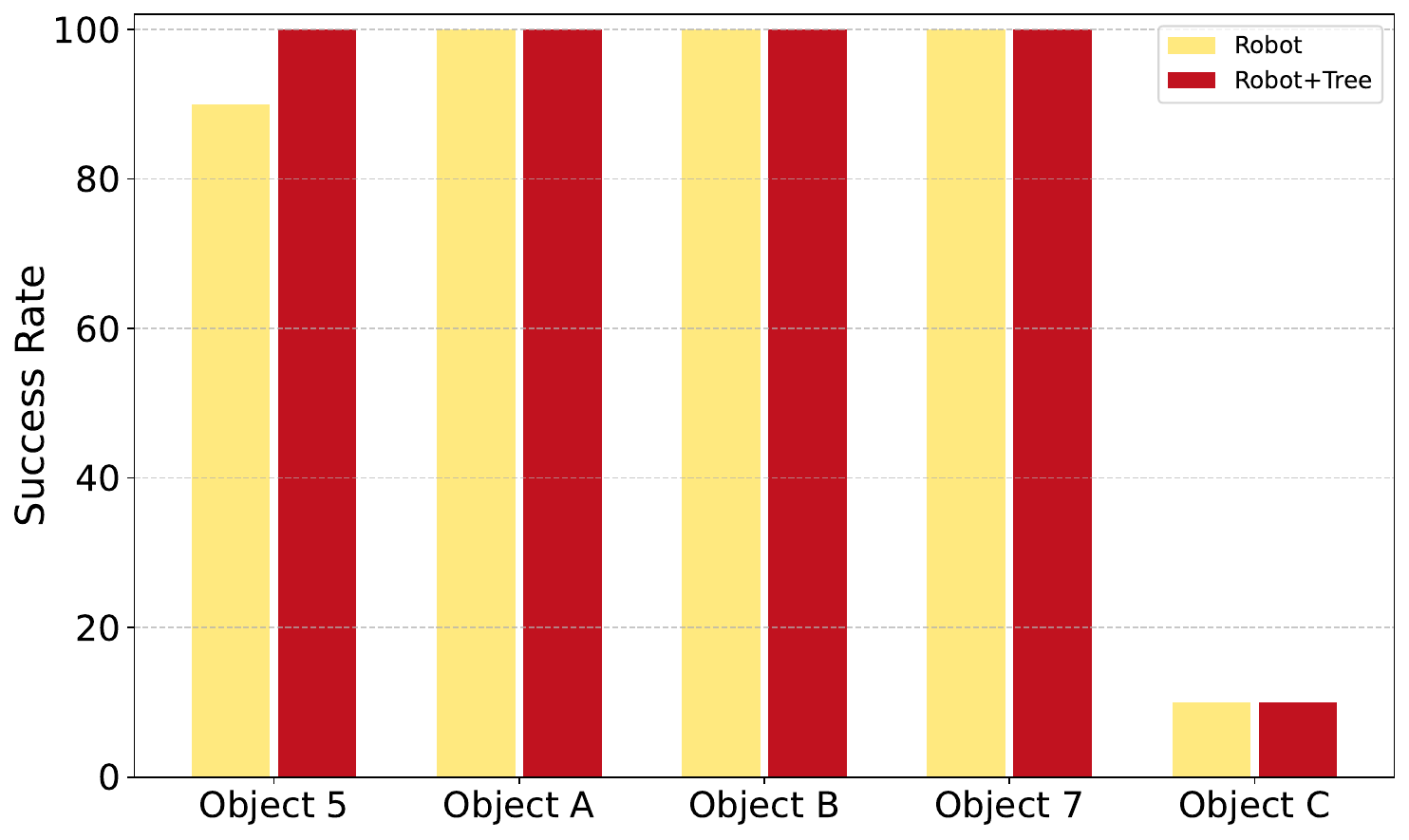}
        \caption{Results of property probing through robotic interaction}
        \label{fig:additional_property_probing_results}
    \end{subfigure}
    \caption{Property probing for similarly shaped objects (e.g., \texttt{white\_3D\_cylinder} and \texttt{yellow\_2D\_rectangle}) exhibiting different physical properties (i.e., compressible, rigid, plastic, foldable). (a) Objects with similar appearances (e.g., Object 5, Object A, Object B; Object 7 and Object C) may possess different physical properties. (b) Robotic interactions provide critical cues that enhance the reasoning capabilities of LLMs.
}
  \label{fig:additional_property_probing}
\end{figure}

To further validate the effectiveness of property probing through robot-object interactions, we conduct additional experiments using objects with similar visual appearances but differing physical properties. 
We select five objects for evaluation, Object 5 and Object 7, along with their visually similar counterparts, Objects A and B for Object 5, and Object C for Object 7, as described in~\ref{fig:additional_property_probing_figs}. The Objects 5, A, and B are labeled \texttt{white\_3D\_cylinder}, while Objects 7 and C are \texttt{yellow\_2D\_rectangle}. Each object is tested in ten trials with both the \texttt{Robot} and \texttt{Robot+Tree} methods.
As shown in Fig.~\ref{fig:additional_property_probing_results}, our method achieve an overall success rate of 82\%. Notably, for 3D objects, the \texttt{Robot} and \texttt{Robot+Tree} methods achieve 96.67\% and 100\% accuracy, respectively. For 2D objects, however, the LLM misclassifies Object C as foldable, a mistake also observed with Object 6 (\texttt{blue\_2D\_rectangle}) in Table~\ref{tab:pp_result}. These results indicates that the thin shape of the 2D objects in side views may obstruct the LLMs to infer their physical properties. The results demonstrate the robustness of our method in identifying the physical properties of visually similar objects, while leaving room for improvement when probing 2D objects.

\subsection{Task Planning and Plan Validation}~\label{sec:exp_TP_PV}

We examine the resulting task plans to see if the plans comply with $\mathcal{C}$ and achieve $S_\text{goal}$. Results of the test instances fall into two cases: (Case I) Constraints are violated and the goal is not reached. (Case II) Constraints are not violated but the goal is not reached. The violation of the constraints can be detected by examining the execution result of the task plan as described in Sec.~\ref{sec:TP_PV}. Also, we can figure out if the goal state is reached by updating and checking the states of objects (i.e., {\texttt{in\_bin}} of all objects are true). Finally, to validate the effectiveness of the discovered object properties in task planning, we also execute several representative task plans with real object.

\begin{figure}[t!]
    \centering
    \begin{subfigure}[b]{0.48\textwidth}
        \captionsetup{skip=0pt}
	    \includegraphics[width=\textwidth]{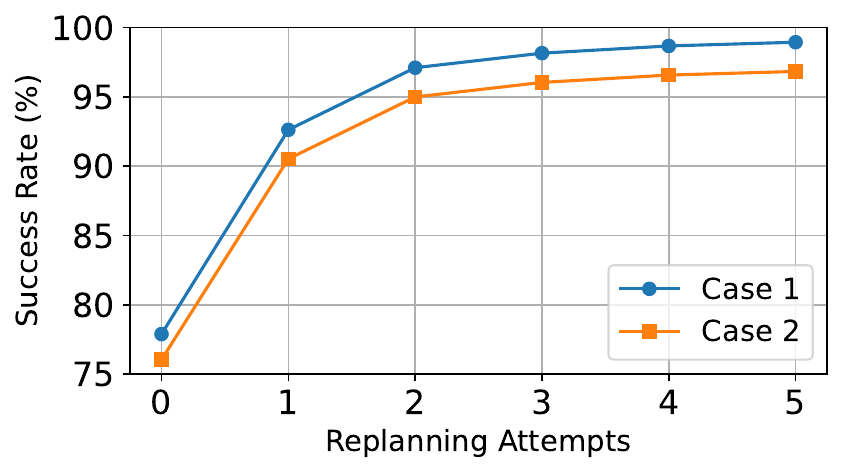}
        \caption{The success rates of task planning and replanning}
        \label{fig:repeat}
    \end{subfigure}
    \quad
    \begin{subfigure}[b]{0.48\textwidth}
        \captionsetup{skip=0pt}
	    \includegraphics[width=\textwidth]{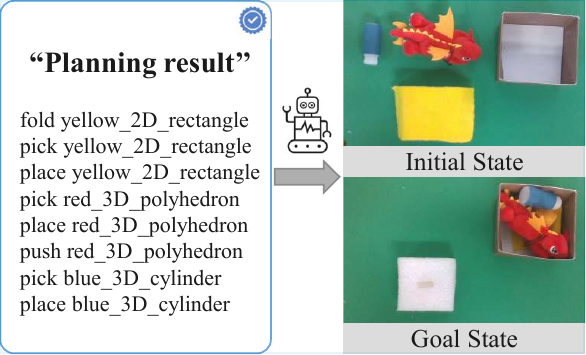}
        \caption{The real-world execution of the task planning}
        \label{fig:task_exe}
    \end{subfigure}
    \caption{The task planning result. (a) The success rates of task planning where replanning has shown the effectiveness of correcting errors. (b) The real-world implementation demonstrates that the probed properties can be effectively utilized in task planning.}
  \label{fig:tp_exp_dum}
\end{figure}

We first measure the success rate of Case I (i.e., violating $\mathcal{C}$ and not reaching $S_\text{goal}$) among 380 instances. As shown in Fig.~\ref{fig:repeat}, 77.89\% of the instances do not violate the constraints in the initial task plan. With one replanning, this rate increases to 92.63\%. After successive four replanning queries, the success rate rises to 97.11\%, 98.16\%, 98.68\%, and 98.95\%, respectively. 
The most significant improvement occurs after the first replanning where most errors are about the constraint violation, which are easy to detect and correct. For example, the sequence of actions {\texttt{pick}}, {\texttt{fold}}, and {\texttt{place}} cannot be executable as the robot cannot fold an object while it is being picked. Another example is applying {\texttt{pick}} action only to rigid objects. The most difficult errors, which are rarely corrected even after a few more replanning queries, are caused by the hallucination of the LLM regarding the constraints. For instance, the constraint {\textit{``Before placing a plastic object, a compressible object should be in the box before''}} often makes the LLM think that one of the objects is plastic even though no plastic object exists.


There are eight instances which belong to Case II. The most common failing instances involve Object 10 ({\texttt{black\allowbreak\_1D\allowbreak\_line}}), where the object is bent but not subsequently picked and placed for packing. However, as the replanning progresses, failures in reaching the goal state significantly decrease. Including both cases, the success rate of initial task planning is 76.05\%. After successive five replanning queries, the success rate improves to 90.53\%, 95.00\%, 96.05\%, 96.58\%, and 96.84\%, respectively.

To illustrate how the discovered object properties can influence the result of task planning, we execute task plan examples as shown in Fig.~\ref{fig:task_exe}.\footnote{Please refer to the~\ref{app:task_planning_demo} and supplemental video for full task planning execution.} All grounded actions in the plan such as ``\texttt{fold yellow\_2D\_rectangle}'' are executed using PbD method. During execution, as the \texttt{yellow\_2D\_rectangle} is known to be foldable, the robot folds the object to utilize bin space for packing. After folding the object, the robot push the \texttt{red\_3D\_polyhedron} after place it into the bin. Finally, the robot pick and place the \texttt{blue\_3D\_cylinder}. This execution provides meaningful validation that the discovered object properties can be effectively integrated into real-world robotic task planning.

\subsection{Discussions}

While the success rate of task planning exceeds 95\% after a few iterations, the first planning query does not give a sufficiently good performance. This limitation might come from the ambiguity in the prompts, limited domain knowledge of the LLM, incomplete understanding of physical or contextual understanding of the LLM, etc. For example, the term `plastic' might provide an ambiguous signal to the LLMs, interpreting it as a material type rather than its intended concept of `plastic deformable' in mechanics. In addition, our method consists of several sequential processes where the result of preceding processes significantly impacts the following processes.

In the experiments, we use the PbD method for the robots to automatically generate motions of probing actions and task execution. Although we do not set out to achieve full autonomy which is not the focus of this work, we attempted to use the state-of-the-art imitation learning method ALOHA~\cite{zhao2023learning}. However, the method did not perform reliably for our probing actions which need delicate and precise actions to handle deformable objects. Specifically, it struggles to recover from early mistakes especially in complex multi-step tasks such as `bend' or `fold' actions. Given these considerations, PbD is a fast and efficient method in scenarios where complex trajectories represent a single action skill while maintaining the consistency of the action. 
We believe that the current level of autonomy we achieved demonstrates the potential of applying our method in the near future where Vision-Language-Action (VLA) models or other advanced learning-based methods can perform tasks handling deformable objects which require precise and robust action generation.

As an initial attempt to understand the physical properties of unseen objects, our work leaves room for improvement in generalization performance, both across diverse object properties and other manipulation tasks. Another direction is to detect irreversible properties without prior knowledge or specialized sensors without causing damage to the object. In the future, we will develop more general probing skills applicable to a wider range of physical properties.

\section{Conclusion and Future Work}~\label{sec:conclusion}
In this paper, we propose an LLM-based reasoning method to understand the physical properties of unseen objects through the interactions between the objects and a robot. Experimental results demonstrate the effectiveness of the proposed method in task planning with iterative improvements. In the future, we further extend our method to a larger set of objects with diverse physical properties. We will also explore various manipulation tasks, such as tidying up homes, which involve handling various fragile and deformable objects. Incorporating additional sensing modalities (e.g., tactile) can help understand the various properties.

\section*{Acknowledgment}

This work was supported by the National Research Foundation of Korea (NRF) grants funded by the Korea government (MSIT) (No. RS-2024-00411007 and No. RS-2024-00461583).

\bibliographystyle{elsarticle-num} 
\bibliography{main}

\newpage
\appendix
\section{Prompts}~\label{app:Prompts}
\setcounter{page}{1}
\subsection{Prompts for Object Detection and Naming (Sec.~\ref{sec:OD})}~\label{app:od}
\begin{tcolorbox}[breakable, title=Prompts for the object detection and naming, title filled=false, colback=black!5!white, colframe=black!75!black, left=2pt, right=2pt, top=2pt, bottom=2pt]
\textbf{Message 1}: The first image shows a top view of an object (the one which is the closest to the center of the image) while the second image shows a side view of them.  \\
Each view captures the same scene, with the objects in the images corresponding one-to-one between the two views. \\
I will provide you with images that have bounding boxes drawn around the objects and its logits.  \\
However, unfortunately, these images can occasionally contain errors.  \\
Therefore, you need to make accurate judgments about the objects' positions and relationships.  \\
Additionally, the white lines represent the relationships between the bounding boxes, so think of this as a graph. \\
 \\
Based on the spatial information from this graph, you should be able to recognize the same objects.  \\
Please describe the given environment based on these images.       \\
 \\
Your answer must use the template below: \\
Please answer with the template below: \\
---template start--- \\
\#\#\# Objects and Their Descriptions: \\
\# top view
\begin{verbatim}
|  top index  | descriptions   
|      1      |
|      2      |
\end{verbatim}
\# side view
\begin{verbatim}
|  side index  | descriptions
|       1      |
|       2      |
\end{verbatim}
\# Find same objects in both views \\
\\
\#\#\# Spatial Relationships:\\
\\
\#\#\# Critical bounding box errors Description: \\
\# if there is any critical error such as wrong bonding box and additional bounding box, describe here  \\
 \\
---template end--- \\

\textbf{Answer 1}:
---template start---\\
\#\#\# Objects and Their Descriptions:\\

\# top view
\begin{verbatim}
|  top index  | descriptions  
|      1      | Brown round container
|      2      | Yellow sponge with a yellow band
|      3      | White cylindrical foam
|      4      | Blue rectangular object
|      5      | Transparent bottle
|      6      | Black cable or wire
\end{verbatim}

\# side view
\begin{verbatim}
|  side index  | descriptions
|       1      | Brown round container
|       2      | Yellow sponge with a yellow band
|       3      | White cylindrical foam
|       4      | Blue rectangular object
|       5      | Transparent bottle
|       6      | Black cable or wire
\end{verbatim}

\# Find same objects in both views \\
- Brown round container: top index 1, side index 1\\
- Yellow sponge with a yellow band: top index 2, side index 2\\
- White cylindrical foam: top index 3, side index 3\\
- Blue rectangular object: top index 4, side index 4\\
- Transparent bottle: top index 5, side index 5\\
- Black cable or wire: top index 6, side index 6\\
\\
\#\#\# Spatial Relationships:\\
- The brown round container is centrally located in both views.\\
- The yellow sponge is to the left of the container in both views.\\
- The white cylindrical foam is to the right of the container in both views.\\
- The blue rectangular object is further right in both views.\\
- The transparent bottle is to the left and slightly below the container in both views.\\
- The black cable or wire is centrally located, below the container in both views.\\
\\
\#\#\# Critical bounding box errors Description:\\
- No critical bounding box errors detected.\\
---template end---\\
\\ 
\textbf{Message 2}: Now that you've obtained the spatial information of the objects, you need to describe shape, color, and dimension of each object. \\
And for clarity, I will provide the original images. Please identify the shape, dimension, and color of each object based on these images according to the definitions in ``[Definitions of dimensions and shapes]" below. \\
Your answer must follow the naming convention which is ``color\_dimension\_shape" (e.g., red\_3D\_cuboid or black\_2D\_ring). \\
Ensure that there is no contradiction between the shape and dimension. For example, ``1D" and ``loop" or ``3D" and ``circle" are not compatible according to their definitions in ``[Definitions of dimensions and shapes]".
\\

[Definitions of dimensions and shapes]

Dimension \\
1D: One-dimensional object, such as thin and sharp thing \\
2D: Two-dimensional object, without height \\
3D: Three-dimensional object \\
 \\
Shape \\
1D \\
Line: A one-dimensional object characterized by a straight path connecting two endpoints, having a definite length but no width. If an object is with an empty center, it should not be a line. \\
 \\
2D \\
Circle: A two-dimensional round object with every point on its edge equidistant from its center \\
Rectangle: A two-dimensional polygonal object with opposite sides equal and four right angles \\
Triangle: A two-dimensional polygonal object with three sides and three angles \\
Polygon: A two-dimensional object with multiple straight sides (e.g., a pentagon is with 5 sides, a hexagon is with 6 sides) \\
Loop: A two-dimensional object with a closed curve characterized by an empty center, forming a continuous loop \\
 \\
3D \\
Sphere: A round three-dimensional object where every point on its surface is equidistant from its center \\
Cube: A three-dimensional object with six equal square faces \\
Cuboid: A three-dimensional object with six rectangular faces \\
Pyramid: A three-dimensional object with a polygonal base and triangular faces that converge to a single point \\
Cylinder: A three-dimensional object with two parallel circular bases connected by a curved surface \\
Cone: A three-dimensional object with a circular base that tapers to a single point \\
Polyhedron: A three-dimensional object with flat polygonal faces, straight edges, and vertices, \\
 \\
Your answer must use the template below: \\
 \\
Please answer with the template below: \\
---template start--- \\
Answer \\
--- \\
object: red\_3D\_polyhedron, yellow\_3D\_cuboid, ... \# if there are duplicate objects, add `\_N' at the end, e.g., red\_3D\_cuboid\_2. \\
--- \\
 \\
Descriptions about objects in the scene \\
*your descriptions in 200 words \\
---template end---
\\
\\
\textbf{Answer 2}:
---template start---\\
\\
Answer\\
---\\
object: brown\_3D\_cylinder, yellow\_3D\_cuboid, white\_3D\_cuboid, blue\_2D\_rectangle, transparent\_3D\_cuboid, black\_1D\_line\\
---\\
Descriptions about objects in the scene:\\
The scene contains six distinct objects. The brown object is a 3D cylinder, likely a container, with a round shape and vertical sides. The yellow object is a 3D cuboid, resembling a sponge with a yellow band around it, giving it a rectangular appearance. The white object is also a 3D cuboid, possibly made of foam, with a rectangular shape. The blue object is a 2D rectangle, flat and rectangular, likely a piece of paper or card. The transparent object is a 3D cuboid, resembling a small bottle with a rectangular shape. Lastly, the black object is a 1D line, thin and elongated, possibly a cable or wire. Each object is distinct in color and shape, contributing to a diverse arrangement in the scene.\\
\\
---template end---\\
\end{tcolorbox}

\subsection{Prompts for Property Probing (Sec.~\ref{sec:property_probing})}~\label{app:pp}
\begin{tcolorbox}[breakable, title=Prompt for the property probing: Robot, title filled=false, colback=black!5!white, colframe=black!75!black, left=2pt, right=2pt, top=2pt, bottom=2pt]
\textbf{Message}: We will now probe the object's properties. \\
This table defines the physical properties of the object we are investigating.\\
is\_rigid: the object is stiff and does not deform easily.\\
is\_bendable: the 1D object can be flexed without breaking or cracking:. \\
is\_foldable: the 2D object can be folded without being damaged.\\
is\_compressible: the 3D object can pressed into a smaller space.\\
is\_plastic: the object can be shaped and retains the new form.\\
\\
\\
!Note1: we do not examine precise physical property of object but for bin\_packing. This mean, when we investigate the properties of an object, we refer to its irreversibility rather than its physical feasibility \\
 \\
object name: red\_3D\_cuboid \\
info \\
We will show you the three images when the robot does the action push on the object.\\
The first image is just before the robot push the object.\\
The second image is the image when the robot push the object.\\
The last image is the image after the robot recovers the object.\\
\\
With these prior knowledge of the object, determine the property of the object. (choose only one property) \\
---template start--- \\
1. Reasoning \\
    -   \\

2. Result \\
    - Property: `' \\
 \\
---template end--- 
\\
\\
\textbf{Answer: }\\
---template start---\\
1. Reasoning\\
    - The object is a 3D cuboid. When the robot pushes the object, it deforms and does not return to its original shape after the force is removed. This indicates that the object can be shaped and retains the new form, which aligns with the property of being plastic.\\
\\
2. Result\\
    - Property: `is\_plastic'\\
---template end---\\

\end{tcolorbox}

\begin{tcolorbox}[breakable, title=Prompt for the property probing: Robot+Tree, title filled=false, colback=black!5!white, colframe=black!75!black, left=2pt, right=2pt, top=2pt, bottom=2pt]
\textbf{Message 1}: Describe the scenes\\
1. \\
2. \\
3. \\
\\
\textbf{Answer 1}: 1. The object is a small, red, rectangular box placed on a flat surface. It appears to be made of cardboard or a similar material. \\
2. The object is the same red box, but it is now being grasped by a black robotic gripper. The box is slightly deformed, indicating pressure from the gripper.\\
\\
\textbf{Message 2}: We will now probe the object's properties. \\
This table defines the physical properties of the object we are investigating. \\
is\_rigid: the object is stiff and does not deform easily. \\
is\_bendable: the 1D object can be flexed without breaking or cracking:.  \\
is\_foldable: the 2D object can be folded without being damaged.
is\_compressible: the 3D object can pressed into a smaller space. \\
is\_plastic: the object can be shaped and retains the new form. \\
\\
 \\
!Note: we do not examine precise physical property of object but for bin\_packing. This mean, when we investigate the properties of an object, we refer to its irreversibility rather than its physical feasibility\\
\\
\# Here are some example of !Note2 \\
\# Ex1: Clothes are foldable, so they can be folded. However, while paper is also foldable (in the meaning of physical feasibility), folding it leaves creases that cannot be undone (The paper is not foldable but plastic in the meaning of irreversibility). \\
\# Ex2: Similarly, while a teddy bear is compressible, a clay toy is not compressible but rather plastic, as it cannot revert back once pushed (in the meaning of irreversibility). \\
\\
In our case, we want to investigate about is\_compressible and is\_rigid. Are you ready? \\
\\
\\
\textbf{Answer 2}: Yes, I'm ready. Based on the images, we can investigate whether the object is compressible or rigid. Please proceed with your observations or questions.\\
\\
\textbf{Message 3}: The images are images of red\_3D\_cuboid. Please refer to this image and answer. \\
We will show you the image when the robot does the action push on the object to verify the object is whether is\_compressible or is\_rigid. \\
The 1st image is just before the robot push the object.  \\
The 2nd image is the image when the robot is push the object.  \\
If the object's shape has changed in the image, consider that `is\_compressible' occurred during push action.Does this object have `is\_compressible' or `is\_rigid' properties?  \\
\\
\\
---template start---\\
1. Description of 1st image:\\
    - \\
    \\
2. Description of 2st image:\\
    - \\
    \\
3. Reasoning\\
    -   \\
     \\
4. Result \\
    - Property: `' \\
\\
---template end---
\\
\\
\textbf{Answer 3-1}: ---template start---\\
1. Description of 1st image:\\
    - The red 3D cuboid is intact and maintains its original shape, placed on a flat surface.\\

2. Description of 2nd image:\\
    - The red 3D cuboid is being pushed by a robotic gripper, and its shape is visibly deformed.\\

3. Reasoning\\
    - The deformation of the cuboid when pushed indicates that it can be compressed, showing a change in shape under pressure.\\

4. Result\\
    - Property: `is\_compressible'\\
---template end---\\
\\
\textbf{Answer 3-2}: ---template start---\\
1. Description of 1st image:\\
    - The red 3D cuboid is intact and centered within a marked area on a flat surface.\\
\\
2. Description of 2nd image:\\
    - A robotic arm is pressing the red cuboid, causing it to deform visibly.\\

3. Reasoning\\
    - After the robot releases the object, the cuboid remains deformed and does not return to its original shape. This indicates that the object does not recover from deformation.\\

4. Result\\
    - Property: `is\_plastic'\\
---template end---
\end{tcolorbox}

\subsection{Prompts for Predicate Generator (Sec.~\ref{sec:DG})}~\label{app:pg}
\begin{tcolorbox}[breakable, title=Prompt for the predicate generator, title filled=false, colback=black!5!white, colframe=black!75!black, left=2pt, right=2pt, top=2pt, bottom=2pt]
\textbf{Message}: Our goal is to define the types of objects and their predicates within the dataclass Object.  \\
Here, we have the types, names, and properties of the objects recognized from the input images. We need to use this information to complete the Object class.  \\
0: name=`white\_3D\_cuboid', predicates=[`is\_rigid'] \\
1: name=`yellow\_3D\_cuboid', predicates=[`is\_compressible'] \\
2: name=`white\_3D\_cylinder', predicates=[`is\_compressible'] \\
3: name=`blue\_2D\_rectangle', predicates=[`is\_plastic'] \\
4: name=`black\_1D\_line', predicates=[`is\_bendable'] \\
5: name=`brown\_3D\_cylinder', predicates=[`is\_compressible']
\begin{verbatim}
from dataclasses import dataclass 

@dataclass 
class Object: 
    # Basic dataclass 
    index: int 
    name: str 
    color: str 
    shape: str 
    object_type: str  # box or obj 
 
    # physical state of an object for bin_packing 
    in_bin: bool 
 
    # Object physical properties  
     
    # Additional predicates for bin_packing (max: 1) 
\end{verbatim}
However, we cannot complete a planning with this dataclass predicate alone which means that we have to add other predicates that fully describe bin\_packing task.  \\
Also you have to add predicates such as is\_rigid, is\_compressible, is\_bendable, and is\_plastic.  \\
Add more predicates needed for bin\_packing to class Object.  \\
 \\
Please answer using the template below: \\
---template start--- \\
Answer: \\
```python \\
\# only write a code here without example instantiation \\
''' \\
Reason: \\
\# Explain in less than 200 words and why you made such predicates. \\
---template end---\\
\\
\textbf{Answer}: 
\begin{verbatim}
from dataclasses import dataclass

@dataclass
class Object:
    # Basic dataclass
    index: int
    name: str
    color: str
    shape: str
    object_type: str  # box or obj

    # physical state of an object for bin_packing
    in_bin: bool

    # Object physical properties
    is_rigid: bool = False
    is_compressible: bool = False
    is_plastic: bool = False
    is_bendable: bool = False

    # Additional predicates for bin_packing (max: 1)
    is_fragile: bool = False


@dataclass
class Box:
    # Basic dataclass
    index: int
    name: str

    # Predicates for box
    object_type: str  # box or obj
    in_bin_objects: list     
\end{verbatim}








\end{tcolorbox}

\subsection{Prompts for Action Generator (Sec.~\ref{sec:DG})}~\label{app:ag}
\begin{tcolorbox}[breakable, title=Prompt for the action generator, title filled=false, colback=black!5!white, colframe=black!75!black, left=2pt, right=2pt, top=2pt, bottom=2pt]
\textbf{Message 1}: Our goal is to define the pre-conditions and effects for the robot's actions, similar to how they are done in PDDL.
\begin{verbatim}
from dataclasses import dataclass 
 
@dataclass 
class Object: 
    # Basic dataclass 
    index: int 
    name: str 
    color: str 
    shape: str 
    object_type: str  # box or obj 
 
    # physical state of an object for bin_packing 
    in_bin: bool 
 
    # Object physical properties 
    is_rigid: bool 
    is_compressible: bool 
    is_bendable: bool 
    is_plastic: bool 
 
    # Additional predicates for bin_packing (max: 1) 
    is_heavy: bool 
 
 
@dataclass 
class Box: 
    # Basic dataclass 
    index: int 
    name: str 
 
    # Predicates for box 
    object_type: str  # box or obj 
    in_bin_objects: list                 
\end{verbatim}

This is the object class we should reference.  \\
What are the `\# Object physical properties' we need to consider? Please provide an answer in 100 words.  \\
1.  \\
2.  \\
3.  \\
... \\
 \\
\textbf{Answer 1}: When considering the physical properties of objects for bin packing, the following properties are essential:
\\
1. **Weight**: Determines if the object is heavy or light, impacting how it can be packed and whether it requires special handling.
\\
2. **Material**: Identifies if the object is plastic, metal, or another material, affecting its fragility and packing requirements.
\\
3. **Rigidity**: Indicates if the object is rigid or flexible, influencing how it can be positioned and stacked.
\\
4. **Size**: Defines the dimensions of the object, crucial for determining fit within a box.
\\
5. **Shape**: Affects how objects can be arranged and packed efficiently.
\\
These properties help in planning and executing effective bin-packing strategies.
\\
 
\textbf{Message 2}: This is Object class you have to use.  
\begin{verbatim}
from dataclasses import dataclass 
 
@dataclass 
class Object: 
    # Basic dataclass 
    index: int 
    name: str 
    color: str 
    shape: str 
    object_type: str  # box or obj 
 
    # physical state of an object for bin_packing 
    in_bin: bool 

    # Object physical properties
    is_rigid: bool
    is_compressible: bool
    is_bendable: bool
    is_plastic: bool

    # Additional predicates for bin_packing (max: 1)
    is_heavy: bool

@dataclass
class Box:
    # Basic dataclass
    index: int
    name: str

    # Predicates for box
    object_type: str  # box or obj
    in_bin_objects: list 
\end{verbatim}
Define the preconditions and effects of actions based on the `Object' class. Use these to create actions. 
\\
However, if a physical property mentioned in the constraints does not exist in the Object class, ignore the constraint and do not reflect it in `def pick(self, obj, bin):'. Generate the function `print(`pick', obj.name)' without making any assumptions.
\begin{verbatim}
class Action: 
    def __init__(self, name: str = "UR5"): 
        self.name = name     
        self.robot_handempty = True 
        self.robot_now_holding = False 
 
    # basic state 
    def state_handempty(self): 
        self.robot_handempty = True 
        self.robot_now_holding = False 
 
    # basic state 
    def state_holding(self, obj): 
        self.robot_handempty = False 
        self.robot_now_holding = obj 
\end{verbatim}
 
Answer:\\
```python
\begin{verbatim}
# only write a `def pick` here without examples and `class Action` 
def pick(self, obj, box): 
    if:  
    # Action Description: Pick an object that not in the box. 
    # The action does not include the `place' action and can be applied to any type of object.    
    # Note! If a predicate required by the constraints is absent from the class Object, 
    # do not consider the constraints. 
        print(f"pick obj.name") 
        # state the effect of Action and box:
        # self.state_holding or self.state_handempty etc if necessary 
        ...effect... 
         
    else: 
        print(f"Cannot pick obj.name")         
\end{verbatim}

'''\\
\textbf{Answer 2}: 
```python
\begin{verbatim}
    def pick(self, obj, box):
        # Preconditions: The object is not in the bin and the robot's hand is empty
        if not obj.in_bin and self.robot_handempty:
            # Action Description: Pick an object that is not in the box. 
            print(f"pick {obj.name}")
            # Effects: The robot is now holding the object, and the hand is not empty
            self.state_holding(obj)
        else:
            print(f"Cannot pick {obj.name}")
\end{verbatim}
'''

\end{tcolorbox}
In this `pick' function, the precondition checks if the object is not already in the box (`not obj.in\_bin') and if the robot's hand is empty (`self.robot\_handempty'). If these conditions are met, the robot picks the object, and the state is updated to reflect that the robot is now holding the object. If the conditions are not met, it prints that the object cannot be picked. Constraints not present in the `Object` class are ignored, as specified.

\subsection{Prompts for Initial State Descriptor (Sec.~\ref{sec:ID})}~\label{app:is}
\begin{tcolorbox}[breakable, title=Prompt for the initial state descriptor, title filled=false, colback=black!5!white, colframe=black!75!black, left=2pt, right=2pt, top=2pt, bottom=2pt]
\textbf{Message}: We are now making initial state of the bin\_packing. We get these information from the input images.  \\
 \\
0: {`name': `white\_3D\_cuboid', `shape': `3D\_cuboid', `color': `white', `predicates': [`is\_rigid'], `init\_pose': `out\_box'},  \\
1: {`name': `yellow\_3D\_cuboid', `shape': `3D\_cuboid', `color': `yellow', `predicates': [`is\_compressible'], `init\_pose': `out\_box'},  \\
2: {`name': `white\_3D\_cylinder', `shape': `3D\_cylinder', `color':`'white', `predicates': [`is\_compressible'], `init\_pose': `out\_box'},  \\
3: {`name': `blue\_2D\_rectangle', `shape': `2D\_rectangle', `color': `blue', `predicates': [`is\_plastic'], `init\_pose': `out\_box'},  \\
4: {`name': `black\_1D\_line', `shape': `1D\_line', `color': `black', `predicates': [`is\_bendable'], `init\_pose': `out\_box'},  \\
5: {`name': `brown\_3D\_cylinder', `shape': `3D\_cylinder', `color': `brown', `predicates': [`is\_compressible'], `init\_pose': `out\_box'}
\begin{verbatim}
from dataclasses import dataclass

@dataclass
class Object:
    # Basic dataclass
    index: int
    name: str
    color: str
    shape: str
    object_type: str  # box or obj

    # physical state of an object for bin_packing
    in_bin: bool

    # Object physical properties
    is_rigid: bool
    is_compressible: bool
    is_bendable: bool
    is_plastic: bool

    # Additional predicates for bin_packing (max: 1)
    is_heavy: bool


@dataclass
class Box:
    # Basic dataclass
    index: int
    name: str

    # Predicates for box
    object_type: str  # box or obj
    in_bin_objects: list                
\end{verbatim}

Using the above information, Please organize the initial state of the domain in a table.
\\
Please answer with the template below: \\
---template start--- \\
\#\#\# 1. Init Table \\
\# fill your table using objects predicates \\
 \\
\#\#\# 2. Python Codes \\
\# make init state into python code \\
\# don't include the object classes or robot class, make only objects and bin  \\
\# example  \\
```python \\
object0 = Object(index=0, name=`black\_3D\_cuboid', color=`black', shape=`3D\_cuboid', ...)\\
object1 = Object(index=1, name=`white\_2D\_circle', color=`white', shape=`2D\_circle', ...)\\
box = Box()\\
''' \\
...\\
\\
\#\#\# 3. Notes:  \\
\# Fill your notes \\
---template end---\\
\\
\textbf{Answer}: ---template start---\\
\#\#\# 1. Init Table
{\footnotesize
\begin{verbatim}
| Index | Name             | .. | .. | In Bin | Is Rigid | Is Compressible | Is Plastic | Is Bendable |
|-------|------------------|----|----|--------|----------|-----------------|------------|-------------|
| 0     | white_3D_cuboid  |    |    | False  | True     | False           | False      | False       |
| 1     | yellow_3D_cuboid |    |    | False  | False    | True            | False      | False       |
| 2     | white_3D_cylinder|    |    | False  | False    | True            | False      | False       |
| 3     | blue_2D_rectangle|    |    | False  | False    | False           | True       | False       |
| 4     | black_1D_line    |    |    | False  | False    | False           | False      | True        |
| 5     | brown_3D_cylinder|    |    | False  | False    | True            | False      | False       |
\end{verbatim}
}

\#\#\# 2. Python Codes

```python
\begin{lstlisting}
object0 = Object(index=0, name="red_3D_cuboid", color="red", shape="3D_cuboid", object_type="obj", in_bin=False, is_plastic=True, is_rigid=False, is_heavy=False)
object1 = Object(index=1, name="white_3D_cuboid", color="white", shape="3D_cuboid", object_type="obj", in_bin=False, is_plastic=False, is_rigid=True, is_heavy=False)
object2 = Object(index=2, name="green_3D_cylinder", color="green", shape="3D_cylinder", object_type="obj", in_bin=False, is_plastic=False, is_rigid=True, is_heavy=False)    
\end{lstlisting}
'''

\#\#\# 3. Notes:

- The initial state indicates that all objects are currently out of the bin.\\
- The predicates `is\_plastic' and `is\_rigid' are used to define the physical properties of the objects.\\
- The `is\_heavy' predicate is not applicable to any of the objects in the initial state.\\
- The `object\_type' is set to `obj' for all items, indicating they are objects to be packed, not boxes.\\
\\
---template end---

\end{tcolorbox}

\subsection{Prompts for Goal State Descriptor (Sec.~\ref{sec:ID})}~\label{app:gs}

\begin{tcolorbox}[breakable, title=Prompt for the goal state descriptor, title filled=false, colback=black!5!white, colframe=black!75!black, left=2pt, right=2pt, top=2pt, bottom=2pt]
\textbf{Message}: We are currently defining the goal state for the bin\_packing. Your task is to translate the goal state provided in natural language into a table. \\
\\
This is a collection of target objects for the bin\_packing task. \\
0: {`name': `white\_3D\_cuboid', `shape': `3D\_cuboid', `color': `white', `predicates': [`is\_rigid'], `init\_pose': `out\_box'},  \\
1: {`name': `yellow\_3D\_cuboid', `shape': `3D\_cuboid', `color': `yellow', `predicates': [`is\_compressible'], `init\_pose': `out\_box'},  \\
2: {`name': `white\_3D\_cylinder', `shape': `3D\_cylinder', `color':`'white', `predicates': [`is\_compressible'], `init\_pose': `out\_box'},  \\
3: {`name': `blue\_2D\_rectangle', `shape': `2D\_rectangle', `color': `blue', `predicates': [`is\_plastic'], `init\_pose': `out\_box'},  \\
4: {`name': `black\_1D\_line', `shape': `1D\_line', `color': `black', `predicates': [`is\_bendable'], `init\_pose': `out\_box'},  \\
5: {`name': `brown\_3D\_cylinder', `shape': `3D\_cylinder', `color': `brown', `predicates': [`is\_compressible'], `init\_pose': `out\_box'} \\
 \\
Our goal is as follows:  \\
Pack all the objects. \\
 \\
Using the information, please organize the goal state of the domain into a table.  \\
 \\
Please answer with the template below: \\
---template start--- \\
\#\#\# 1. Goal Table
\begin{verbatim}
| Index | Name | State1 | State2 | ...
|-------|------|--------|--------|-----
|   0   |      |  True  |  False | ...
|   1   |      |  True  |  True  | ...
|   2   |      |  False |  False | ...
|  ...  |  ..  |   ..   |   ..   | ...
\end{verbatim}

\#\#\# 2. Notes: \\
\# Fill your notes \\
---template end--- \\

\textbf{Answer}: ---template start---\\
\#\#\# 1. Goal Table \\
\begin{verbatim}
| Index | Name              | Packed |
|-------|-------------------|--------|
|   0   | white_3D_cuboid   |  True  |
|   1   | yellow_3D_cuboid  |  True  |
|   2   | white_3D_cylinder |  True  |
|   3   | blue_2D_rectangle |  True  |
|   4   | black_1D_line     |  True  |
|   5   | brown_3D_cylinder |  True  |
\end{verbatim}

\#\#\# 2. Notes:\\
- The goal state for the bin\_packing task is to have all objects packed.\\
- Each object is initially out of the box and needs to be packed to achieve the goal state.\\
- The table reflects the desired state where all objects are packed, indicated by ``True'' under the ``Packed'' column.\\
---template end---
\end{tcolorbox}

\subsection{Prompts for Task Planner (Sec.~\ref{sec:TP_PV})}~\label{app:tp}
\begin{tcolorbox}[breakable, title=Prompt for the task planner, title filled=false, colback=black!5!white, colframe=black!75!black, left=2pt, right=2pt, top=2pt, bottom=2pt]
\textbf{Message 1}: Refer the following code containing the list of actions (class Action) and the set of initial states of objects which are defined in class Object. 
\begin{lstlisting}
from dataclasses import dataclass

@dataclass
class Object:
    # Basic dataclass
    index: int
    name: str
    color: str
    shape: str
    object_type: str  # box or obj

    # physical state of an object for bin_packing
    in_bin: bool

    # Object physical properties
    is_rigid: bool
    is_compressible: bool
    is_bendable: bool
    is_plastic: bool

    # Additional predicates for bin_packing (max: 1)
    is_heavy: bool


@dataclass
class Box:
    # Basic dataclass
    index: int
    name: str

    # Predicates for box
    object_type: str  # box or obj
    in_bin_objects: list                



class Action:
    def __init__(self, name: str = "UR5"):
        self.name = name    
        self.robot_handempty = True
        self.robot_now_holding = False

    # basic state
    def state_handempty(self):
        self.robot_handempty = True
        self.robot_now_holding = False

    # basic state
    def state_holding(self, obj):
        self.robot_handempty = False
        self.robot_now_holding = obj

    def pick(self, obj, box):
        # Preconditions: The robot's hand must be empty, and the object must not already be in the box.
        if self.robot_handempty and not obj.in_bin:
            # Action Description: Pick an object that is not in the box. The action does not include the 'place' action and can be applied to any type of object.
            print(f"pick {obj.name}")
            # Effects: The robot is now holding the object, and the hand is no longer empty.
            self.state_holding(obj)
        else:
            print(f"Cannot pick {obj.name}")

    def place(self, obj, box):
        # Check if the object is plastic and if a compressible object is already in the box
        if obj.is_plastic:
            compressible_in_box = any(o.is_compressible for o in box.in_bin_objects)
            if not compressible_in_box:
                print(f"Cannot place {obj.name}")
                return
        
        # Action Description: Place an object into the box. This action can be applied to any type of object.
        print(f"place {obj.name}")
        
        # Update the state to reflect the object is now in the box
        obj.in_bin = True
        box.in_bin_objects.append(obj)
        
        # Update the robot's state to hand empty after placing the object
        self.state_handempty()

    def bend(self, obj, box):
        # Check if the object is bendable and not plastic
        if obj.is_bendable and not obj.is_plastic:
            # Action Description: Bend a bendable object. Hand must remain empty before and after the bending action.
            print(f"bend {obj.name}")
            # Assuming the robot's hand must be empty before and after bending
            self.state_handempty()
            # Effect: Object is now bent, but since we don't have a specific state for 'bent', we assume it's ready for packing
            # No change in the object's in_bin state here, as bending doesn't place it in the bin
        else:
            print(f"Cannot bend {obj.name}")

    def push(self, obj, box):
        # Action Description: Push a 3D compressible object downward in the bin. 
        # Hand must remain empty before and after the pushing.
        
        # Check preconditions: object must be compressible and hand must be empty
        if obj.is_compressible and self.robot_handempty:
            print(f"push {obj.name}")
            # Effect: Hand remains empty after pushing
            self.state_handempty()
        else:
            print(f"Cannot push {obj.name}")

    def fold(self, obj, box):
        print("Cannot fold")


object0 = Object(index=0, name="white_3D_cuboid", color="white", shape="3D_cuboid", object_type="obj", in_bin=False, is_rigid=True, is_compressible=False, is_bendable=False, is_plastic=False, is_heavy=False)
object1 = Object(index=1, name="yellow_3D_cuboid", color="yellow", shape="3D_cuboid", object_type="obj", in_bin=False, is_rigid=False, is_compressible=True, is_bendable=False, is_plastic=False, is_heavy=False)
object2 = Object(index=2, name="white_3D_cylinder", color="white", shape="3D_cylinder", object_type="obj", in_bin=False, is_rigid=False, is_compressible=True, is_bendable=False, is_plastic=False, is_heavy=False)
object3 = Object(index=3, name="blue_2D_rectangle", color="blue", shape="2D_rectangle", object_type="obj", in_bin=False, is_rigid=False, is_compressible=False, is_bendable=False, is_plastic=True, is_heavy=False)
object4 = Object(index=4, name="black_1D_line", color="black", shape="1D_line", object_type="obj", in_bin=False, is_rigid=False, is_compressible=False, is_bendable=True, is_plastic=False, is_heavy=False)
object5 = Object(index=5, name="brown_3D_cylinder", color="brown", shape="3D_cylinder", object_type="obj", in_bin=False, is_rigid=False, is_compressible=True, is_bendable=False, is_plastic=False, is_heavy=False)

\end{lstlisting}

And this is the table of goal states where all objects should reach. 
\begin{verbatim}
| Index | Name             | Packed |
|-------|------------------|--------|
|   0   | white_3D_cuboid  |  True  |
|   1   | yellow_3D_cuboid |  True  |
|   2   | white_3D_cylinder|  True  |
|   3   | blue_2D_rectangle|  True  |
|   4   | black_1D_line    |  True  |
|   5   | brown_3D_cylinder|  True  |
\end{verbatim}
Fully understand the actions in the Action class and the initial states of all objects. Then tell me your understanding in 300 words. \\

\textbf{Answer 1}: 
The provided code defines a system for managing objects and actions related to a bin-packing scenario. The `Object' class represents various items with attributes like color, shape, and physical properties such as rigidity, compressibility, plasticity, and bendability. Each object also has a boolean `in\_bin' attribute indicating whether it is packed in a bin. The `Box' class represents a container with a list of objects it holds.\\

The `Action' class models a robotic system capable of manipulating these objects. It includes methods for picking, placing, bending, pushing, and folding objects, with specific preconditions and effects for each action. The robot can only pick an object if it is not already in the bin and the robot's hand is empty. Once picked, the robot holds the object, making its hand no longer empty. The `place' method allows the robot to put an object into a box, with constraints such as not placing a plastic object if a compressible object is already present in the box. The `bend' action is applicable to bendable, non-plastic objects, requiring the robot's hand to be empty before and after the action. The `push' action is designed for compressible objects, again requiring the robot's hand to be empty.\\

The initial states of the objects indicate that none are packed in a bin. The goal is to have all objects packed, as specified in the goal state table. The robot must navigate the constraints and properties of each object to achieve this goal. For instance, the robot must handle compressible and plastic objects carefully due to their specific constraints. The system's design emphasizes the importance of understanding object properties and the sequence of actions to achieve the desired packed state for all objects. This setup provides a framework for simulating and solving bin-packing problems with a robotic agent, highlighting the complexities of physical interactions and constraints in automated systems.\\

\textbf{Message 2}: Before start, you must follow the rules: \\
1: Before placing a plastic object, a compressible object should be in the box before. Non-plastic objects can be placed without any constraints. \\
2: Push all compressible objects after placing them in the box. \\
3: If there is a bendalbe object, bend it before placing it in the box. \\
4: If there is a foldable object, fold it before placing it in the box. \\
5: Do not bend, fold, or push a plastic object. \\
\\
Also, we remind you the robot actions: \\
pick: Pick an object that not in the box. The action does not include the `place' action and can be applied to any type of object. \\
place: Place an object into the box. This action can be applied to any type of object. \\
bend: Bend a 1D bendable object. Hand must remain empty before and after the bending action. \\
push: Push a 3D compressible object downward in the bin. Hand must remain empty before and after the pushing. \\
fold: Fold a 2D foldable object. Hand must remain empty before and after the folding. \\
\\
Please answer with the template below: \\

\textbf{Answer 2}: 
```python
\begin{lstlisting}
if __name__ == "__main__":
    # First, from initial state, recall the physical properties of objects and available actions:
    # object0.is_rigid is True, actions pick, place are applicable
    # object1.is_compressible is True, actions pick, place, push are applicable
    # object2.is_compressible is True, actions pick, place, push are applicable
    # object3.is_plastic is True, actions pick, place are applicable
    # object4.is_bendable is True, actions pick, place, bend are applicable
    # object5.is_compressible is True, actions pick, place, push are applicable
    
    # Rewrite the goal states of all objects given in the table in the following format.
    # object0: in_bin: True
    # object1: in_bin: True
    # object2: in_bin: True
    # object3: in_bin: True
    # object4: in_bin: True
    # object5: in_bin: True

    # Second, write a bin_packing order based on the given rules and the goal states of the objects.
    # 1. Bend object4 (black_1D_line) before placing it.
    # 2. Pick and place object4.
    # 3. Pick and place object1 (yellow_3D_cuboid), then push it.
    # 4. Pick and place object2 (white_3D_cylinder), then push it.
    # 5. Pick and place object5 (brown_3D_cylinder), then push it.
    # 6. Pick and place object3 (blue_2D_rectangle) after a compressible object is in the box.
    # 7. Pick and place object0 (white_3D_cuboid).

    # Third, make an action sequence.
    # a) Initialize the robot and the box
    action = Action() 
    box = Box(index=0, name="box1", object_type="box", in_bin_objects=[])

    # b) Action sequence
    action.bend(object4, box)
    action.pick(object4, box)
    action.place(object4, box)

    action.pick(object1, box)
    action.place(object1, box)
    action.push(object1, box)

    action.pick(object2, box)
    action.place(object2, box)
    action.push(object2, box)

    action.pick(object5, box)
    action.place(object5, box)
    action.push(object5, box)

    action.pick(object3, box)
    action.place(object3, box)

    action.pick(object0, box)
    action.place(object0, box)

    # Fourth, after making all actions, provide your reasoning based on the given rules.
    # The sequence follows the rules: bend the bendable object before placing it, ensure a compressible object is in the box before placing the plastic object, and push all compressible objects after placing them. The sequence ensures all objects are placed in the box according to their properties and constraints.

    # Finally, add this code    
    print("All task planning is done")
\end{lstlisting}
'''

\end{tcolorbox}

\subsection{Prompts for Replanning (Sec.~\ref{sec:TP_PV})}~\label{app:rp}
\begin{tcolorbox}[breakable, title=Prompt for the Replanning, title filled=false, colback=black!5!white, colframe=black!75!black, left=2pt, right=2pt, top=2pt, bottom=2pt]
\textbf{Message 1}: This is a python code that you are going to fix.
\begin{lstlisting}
from dataclasses import dataclass

@dataclass
class Object:
    # Basic dataclass
    index: int
    name: str
    color: str
    shape: str
    object_type: str  # box or obj

    # physical state of an object for bin_packing
    in_bin: bool

    # Object physical properties 
    is_rigid: bool = False
    is_compressible: bool = False
    is_bendable: bool = False
    is_plastic: bool = False

    # Additional predicates for bin_packing (max: 1)
    is_fragile: bool = False


@dataclass
class Box:
    # Basic dataclass
    index: int
    name: str

    # Predicates for box
    object_type: str  # box or obj
    in_bin_objects: list

class Action:
    def __init__(self, name: str = "UR5"):
        self.name = name    
        self.robot_handempty = True
        self.robot_now_holding = False

    # basic state
    def state_handempty(self):
        self.robot_handempty = True
        self.robot_now_holding = False

    # basic state
    def state_holding(self, obj):
        self.robot_handempty = False
        self.robot_now_holding = obj

    def pick(self, obj, box):
        # Preconditions: The object must not be in the bin, and the robot's hand must be empty.
        if not obj.in_bin and self.robot_handempty:
            # Action Description: Pick an object that is not in the box.
            print(f"pick {obj.name}")
            # Effects: The robot is now holding the object, and its hand is no longer empty.
            self.state_holding(obj)
        else:
            print(f"Cannot pick {obj.name}")

    def place(self, obj, box):
        # Action Description: Place an object into the box. This action can be applied to any type of object.
        # Check if the object is plastic and if the constraint is applicable
        if obj.is_plastic:
            # Check if there's a compressible object already in the box
            if any(o.is_compressible for o in box.in_bin_objects):
                print(f"place {obj.name}")
                self.state_handempty()  # Assuming placing the object means the robot hand is now empty
                box.in_bin_objects.append(obj)  # Add the object to the box
            else:
                print(f"Cannot place {obj.name} because no compressible object is in the box")
        else:
            # Non-plastic objects can be placed without constraints
            print(f"place {obj.name}")
            self.state_handempty()  # Assuming placing the object means the robot hand is now empty
            box.in_bin_objects.append(obj)  # Add the object to the box

    def bend(self, obj, box):
        # Action Description: Bend a 1D bendable object. Hand must remain empty before and after the bending action.
        # Note! If a predicate required by the constraints is absent from the class Object, do not consider the constraints.
        
        # Check if the object is bendable and the robot's hand is empty
        if obj.is_bendable and self.robot_handempty:
            print(f"bend {obj.name}")
            # State the effect of Action: Hand remains empty after bending
            self.state_handempty()
        else:
            print(f"Cannot bend {obj.name}")

    def push(self, obj, box):
        # Preconditions: The object must be compressible and in the bin, and the robot's hand must be empty.
        if obj.is_compressible and obj.in_bin and self.robot_handempty:
            # Action Description: Push a 3D compressible object downward in the bin. 
            # Hand must remain empty before and after the pushing.
            print(f"push {obj.name}")
            # Effects: The robot's hand remains empty after the action.
            self.state_handempty()
        else:
            print(f"Cannot push {obj.name}")

    def fold(self, obj, box):
        print("Cannot fold")

    def dummy(self):
        pass

# Object Initial State
object0 = Object(index=0, name="white_3D_cuboid", color="white", shape="3D_cuboid", object_type="obj", in_bin=False, is_rigid=True)
object1 = Object(index=1, name="yellow_3D_cuboid", color="yellow", shape="3D_cuboid", object_type="obj", in_bin=False, is_compressible=True)
object2 = Object(index=2, name="white_3D_cylinder", color="white", shape="3D_cylinder", object_type="obj", in_bin=False, is_compressible=True)
object3 = Object(index=3, name="blue_2D_rectangle", color="blue", shape="2D_rectangle", object_type="obj", in_bin=False, is_plastic=True)
object4 = Object(index=4, name="black_1D_line", color="black", shape="1D_line", object_type="obj", in_bin=False, is_bendable=True)
object5 = Object(index=5, name="brown_3D_cylinder", color="brown", shape="3D_cylinder", object_type="obj", in_bin=False, is_compressible=True)

if __name__ == "__main__":
    # First, from initial state, recall the physical properties of objects and available actions:
    # object0.is_rigid is True, actions pick, place are applicable
    # object1.is_compressible is True, actions pick, place, push are applicable
    # object2.is_compressible is True, actions pick, place, push are applicable
    # object3.is_plastic is True, actions pick, place are applicable (with constraints)
    # object4.is_bendable is True, actions pick, bend, place are applicable
    # object5.is_compressible is True, actions pick, place, push are applicable

    # Rewrite the goal states of all objects given in the table in the following format.
    # object0: in_bin: True
    # object1: in_bin: True
    # object2: in_bin: True
    # object3: in_bin: True
    # object4: in_bin: True
    # object5: in_bin: True

    # Second, write a bin_packing order based on the given rules and the goal states of the objects.
    # 1. Bend object4 (black_1D_line) before placing it.
    # 2. Pick and place object1 (yellow_3D_cuboid) and push it.
    # 3. Pick and place object2 (white_3D_cylinder) and push it.
    # 4. Pick and place object5 (brown_3D_cylinder) and push it.
    # 5. Pick and place object3 (blue_2D_rectangle) after a compressible object is in the box.
    # 6. Pick and place object0 (white_3D_cuboid).

    # Third, make an action sequence.
    # a) Initialize the robot and the box
    action = Action()
    box = Box(index=0, name="box1", object_type="box", in_bin_objects=[])

    # b) Action sequence
    action.bend(object4, box)  # Bend the bendable object
    action.pick(object4, box)
    action.place(object4, box)

    action.pick(object1, box)
    action.place(object1, box)
    action.push(object1, box)

    action.pick(object2, box)
    action.place(object2, box)
    action.push(object2, box)

    action.pick(object5, box)
    action.place(object5, box)
    action.push(object5, box)

    action.pick(object3, box)
    action.place(object3, box)

    action.pick(object0, box)
    action.place(object0, box)

    # Fourth, after making all actions, provide your reasoning based on the given rules.
    # The sequence follows the rules: bend the bendable object first, place a compressible object before a plastic one,
    # and push all compressible objects after placing them. The sequence ensures all objects are packed according to their properties.

    # Finally, add this code    
    print("All task planning is done")
\end{lstlisting}

This code consists of four parts below. \\
1. Object Class (Start with [@dataclass]) \\
2. Action Class (Start with [class Action:]) \\
3. Object initial State (Start with [\# Object Initial State]) \\
4. Planning State (Start with [if \_\_name\_\_ == "\_\_main\_\_":]) \\

\ 

And this is a planning result of the code. \\
\texttt{bend black\_1D\_line \\
pick black\_1D\_line \\
place black\_1D\_line \\
pick yellow\_3D\_cuboid \\
place yellow\_3D\_cuboid \\
Cannot push yellow\_3D\_cuboid \\
pick white\_3D\_cylinder \\
place white\_3D\_cylinder \\
Cannot push white\_3D\_cylinder \\
pick brown\_3D\_cylinder \\
place brown\_3D\_cylinder \\
Cannot push brown\_3D\_cylinder \\
pick blue\_2D\_rectangle \\
place blue\_2D\_rectangle \\
pick white\_3D\_cuboid \\
place white\_3D\_cuboid\\
All task planning is done\\}
 \\
Here are constraints you should refer. \\
rules:  \\
`1': `Before placing a plastic object, a compressible object should be in the box before. Non-plastic objects can be placed without any constraints. ' \\
`2': `Push all compressible objects after placing them in the box. ' \\
`3': `If there is a bendalbe object, bend it before placing it in the box. ' \\
`4': `If there is a foldable object, fold it before placing it in the box. ' \\
`5': `Do not bend, fold, or push a plastic object. ' \\
 \\
Your task is to identify any errors in the preconditions and effects of the action within the given Action class. \\
!!! One notable thing is that, in many cases, these actions may be error-free.  \\
Therefore, if no errors are detected, simply return it as is. \\
 \\
Please answer with the template below: \\
---template start--- \\
First, analyze the error message. \\
1.\\
2.\\
3.\\
\\
Second, check the error part follows the notes \\
!!Note1. Do not assume the physical properties of the object. \\
!!Note2. If a predicate required by the constraints is not defined in the class Object, ignore the constraints please.  \\
\# For example, treating a rigid object as a compressible object. They are different. \\
\# For example, treating a bendable object as a foldable object. They are different. \\
 \\
Third, find the error part of the class Action \\
\# Example \\
\# def place() part is wrong. In our objects set, there is no compressible object.  \\
\# However, the precondtion in `def place()' violates the Note2 because they make precondtions ~~~ \\
 \\
---template end---\\
\\
\textbf{Answer 1}: 
---template start---

First, analyze the error message.
1. The error message indicates that the `push' action for compressible objects (yellow\_3D\_cuboid, white\_3D\_cylinder, brown\_3D\_cylinder) is not being executed successfully, as indicated by "Cannot push".\\
2. The `place' action for the blue\_2D\_rectangle (plastic object) is executed without any issue, which aligns with the constraints.\\
3. The sequence of actions seems to follow the given rules, but the `push' action fails for compressible objects.\\

Second, check the error part follows the notes\\
!!Note1. Do not assume the physical properties of the object.\\
!!Note2. If a predicate required by the constraints is not defined in the class Object, ignore the constraints please. \\
\# For example, treating a rigid object as a compressible object. They are different.\\
\# For example, treating a bendable object as a foldable object. They are different.\\

Third, find the error part of the class Action\\
\# The `push' method is incorrect. The precondition for `push` requires the object to be in the bin (`obj.in\_bin'), but this attribute is not updated when the object is placed in the box. \\
\# The `place' method should update `obj.in\_bin' to `True' after placing the object in the box.\\
\# Correct the `place' method to include `obj.in\_bin = True' after `box.in\_bin\_objects.append(obj)' to ensure the `push' action can be executed.\\
\\
---template end---\\

\textbf{Message 2}: \textit{... omitted as the Python code is identical to that described in App.~\ref{app:tp}.}\\

\end{tcolorbox}

\section{Definitions of Shapes, Dimensions, and Properties (Sec.~\ref{sec:OD})}~\label{app:Def}
\subsection{Shapes}
\begin{itemize}
    \item 1D
        \begin{itemize}
            \item Line: A one-dimensional object characterized by a straight path connecting two endpoints, having a definite length but no width. If an object is with an empty center, it should not be a line.
        \end{itemize}
    \item 2D
        \begin{itemize}
            \item Circle: A two-dimensional round object with every point on its edge equidistant from its center
            \item Rectangle: A two-dimensional polygonal object with opposite sides equal and four right angles
            \item Triangle: A two-dimensional polygonal object with three sides and three angles.
            \item Polygon: A two-dimensional object with multiple straight sides (e.g., a pentagon is with 5 sides, a hexagon is with 6 sides)
            \item Loop: A two-dimensional object with a closed curve characterized by an empty center, forming a continuous loop
        \end{itemize}
    \item 3D
        \begin{itemize}
            \item Sphere: A round three-dimensional object where every point on its surface is equidistant from its center
            \item Cube: A three-dimensional object with six equal square faces
            \item Cuboid: A three-dimensional object with six rectangular faces
            \item Pyramid: A three-dimensional object with a polygonal base and triangular faces that converge to a single point
            \item Cylinder: A three-dimensional object with two parallel circular bases connected by a curved surface
            \item Cone: A three-dimensional object with a circular base that tapers to a single point
            \item Polyhedron: A three-dimensional object with flat polygonal faces, straight edges, and vertices
        \end{itemize}
\end{itemize}

\subsection{Dimensions}

\begin{itemize}
    \item 1D: One-dimensional object, such as thin and sharp thing
    \item 2D: Two-dimensional object, without height
    \item 3D: Three-dimensional object 
\end{itemize}

\subsection{Properties}
\begin{itemize}
    \item Rigid: the object is stiff and does not deform easily
    \item Bendable: the 1D object can be flexed without breaking or cracking
    \item Foldable: the 2D object can be folded without being damaged
    \item Compressible: the 3D object can pressed into a smaller space
    \item Plastic: the object can be shaped and retains the new form
\end{itemize}

\newpage
\section{Results of Task Planner and Plan Validator (Sec.~\ref{sec:exp_TP_PV})}

\subsection{Task Planning}~\label{app:TP_Result}
\small\begin{lstlisting}
from dataclasses import dataclass

@dataclass
class Object:
    # Basic dataclass
    index: int
    name: str
    color: str
    shape: str
    object_type: str  # box or obj

    # physical state of an object for bin_packing
    in_bin: bool

    # Object physical properties
    is_rigid: bool
    is_compressible: bool
    is_bendable: bool
    is_plastic: bool

    # Additional predicates for bin_packing (max: 1)
    is_heavy: bool

@dataclass
class Box:
    # Basic dataclass
    index: int
    name: str

    # Predicates for box
    object_type: str  # box or obj
    in_bin_objects: list                

class Action:
    def __init__(self, name: str = "UR5"):
        self.name = name    
        self.robot_handempty = True
        self.robot_now_holding = False

    # basic state
    def state_handempty(self):
        self.robot_handempty = True
        self.robot_now_holding = False

    # basic state
    def state_holding(self, obj):
        self.robot_handempty = False
        self.robot_now_holding = obj

    def pick(self, obj, box):
        # Preconditions: The robot's hand must be empty, and the object must not already be in the box.
        if self.robot_handempty and not obj.in_bin:
            # Action Description: Pick an object that is not in the box. The action does not include the `place' action and can be applied to any type of object.
            print(f"pick {obj.name}")
            # Effects: The robot is now holding the object, and the hand is no longer empty.
            self.state_holding(obj)
        else:
            print(f"Cannot pick {obj.name}")

    def place(self, obj, box):
        # Check if the object is plastic and if a compressible object is already in the box
        if obj.is_plastic:
            compressible_in_box = any(o.is_compressible for o in box.in_bin_objects)
            if not compressible_in_box:
                print(f"Cannot place {obj.name}")
                return
        
        # Action Description: Place an object into the box. This action can be applied to any type of object.
        print(f"place {obj.name}")
        
        # Update the state to reflect the object is now in the box
        obj.in_bin = True
        box.in_bin_objects.append(obj)
        
        # Update the robot's state to hand empty after placing the object
        self.state_handempty()

    def bend(self, obj, box):
        # Check if the object is bendable and not plastic
        if obj.is_bendable and not obj.is_plastic:
            # Action Description: Bend a bendable object. Hand must remain empty before and after the bending action.
            print(f"bend {obj.name}")
            # Assuming the robot's hand must be empty before and after bending
            self.state_handempty()
            # Effect: Object is now bent, but since we don't have a specific state for `bent', we assume it's ready for packing
            # No change in the object's in_bin state here, as bending doesn't place it in the bin
        else:
            print(f"Cannot bend {obj.name}")

    def push(self, obj, box):
        # Action Description: Push a 3D compressible object downward in the bin. 
        # Hand must remain empty before and after the pushing.
        
        # Check preconditions: object must be compressible and hand must be empty
        if obj.is_compressible and self.robot_handempty:
            print(f"push {obj.name}")
            # Effect: Hand remains empty after pushing
            self.state_handempty()
        else:
            print(f"Cannot push {obj.name}")

    def fold(self, obj, box):
        print("Cannot fold")

    def dummy(self):
        pass

# Object Initial State
object0 = Object(index=0, name="white_3D_cuboid", color="white", shape="3D_cuboid", object_type="obj", in_bin=False, is_rigid=True, is_compressible=False, is_bendable=False, is_plastic=False, is_heavy=False)
object1 = Object(index=1, name="yellow_3D_cuboid", color="yellow", shape="3D_cuboid", object_type="obj", in_bin=False, is_rigid=False, is_compressible=True, is_bendable=False, is_plastic=False, is_heavy=False)
object2 = Object(index=2, name="white_3D_cylinder", color="white", shape="3D_cylinder", object_type="obj", in_bin=False, is_rigid=False, is_compressible=True, is_bendable=False, is_plastic=False, is_heavy=False)
object3 = Object(index=3, name="blue_2D_rectangle", color="blue", shape="2D_rectangle", object_type="obj", in_bin=False, is_rigid=False, is_compressible=False, is_bendable=False, is_plastic=True, is_heavy=False)
object4 = Object(index=4, name="black_1D_line", color="black", shape="1D_line", object_type="obj", in_bin=False, is_rigid=False, is_compressible=False, is_bendable=True, is_plastic=False, is_heavy=False)
object5 = Object(index=5, name="brown_3D_cylinder", color="brown", shape="3D_cylinder", object_type="obj", in_bin=False, is_rigid=False, is_compressible=True, is_bendable=False, is_plastic=False, is_heavy=False)


if __name__ == "__main__":
    # First, from initial state, recall the physical properties of objects and available actions:
    # object0.is_rigid is True, actions pick, place are applicable
    # object1.is_compressible is True, actions pick, place, push are applicable
    # object2.is_compressible is True, actions pick, place, push are applicable
    # object3.is_plastic is True, actions pick, place are applicable
    # object4.is_bendable is True, actions pick, place, bend are applicable
    # object5.is_compressible is True, actions pick, place, push are applicable
    
    # Rewrite the goal states of all objects given in the table in the following format.
    # object0: in_bin: True
    # object1: in_bin: True
    # object2: in_bin: True
    # object3: in_bin: True
    # object4: in_bin: True
    # object5: in_bin: True

    # Second, write a bin_packing order based on the given rules and the goal states of the objects.
    # 1. Bend object4 (black_1D_line) before placing it.
    # 2. Pick and place object1 (yellow_3D_cuboid) and push it.
    # 3. Pick and place object5 (brown_3D_cylinder) and push it.
    # 4. Pick and place object2 (white_3D_cylinder) and push it.
    # 5. Pick and place object3 (blue_2D_rectangle) after a compressible object is in the box.
    # 6. Pick and place object0 (white_3D_cuboid).

    # Third, make an action sequence.
    # a) Initialize the robot and the box
    action = Action() 
    box = Box(index=0, name="box", object_type="box", in_bin_objects=[])

    # b) Action sequence
    action.bend(object4, box)
    action.pick(object4, box)
    action.place(object4, box)

    action.pick(object1, box)
    action.place(object1, box)
    action.push(object1, box)

    action.pick(object5, box)
    action.place(object5, box)
    action.push(object5, box)

    action.pick(object2, box)
    action.place(object2, box)
    action.push(object2, box)

    action.pick(object3, box)
    action.place(object3, box)

    action.pick(object0, box)
    action.place(object0, box)

    # Fourth, after making all actions, provide your reasoning based on the given rules.
    # The sequence follows the rules: bend the bendable object before placing, place a compressible object before a plastic one, and push compressible objects after placing them. The sequence ensures that all objects are placed in the box according to their properties and the given constraints.

    # Finally, add this code    
    print("All task planning is done")
\end{lstlisting}

\subsection{Plan Validation}~\label{app:PV_Result}

An example result of the plan validator when task planning is not successful: \\
\texttt{\small pick white\_3D\_cuboid, place white\_3D\_cuboid, pick blue\_3D\_cylinder \\
place blue\_3D\_cylinder, pick blue\_2D\_rectangle, Cannot place blue\_2D\_rectangle \\
Cannot pick gray\_1D\_line, Cannot place gray\_1D\_line}\\

An example result of the plan validator when task planning is successful: \\
\texttt{\small bend black\_1D\_line, pick black\_1D\_line, place black\_1D\_line, pick yellow\_3D\_cuboid\\
place yellow\_3D\_cuboid, push yellow\_3D\_cuboid, pick brown\_3D\_cylinder\\
place brown\_3D\_cylinder, push brown\_3D\_cylinder, pick white\_3D\_cylinder\\
place white\_3D\_cylinder, push white\_3D\_cylinder, pick blue\_2D\_rectangle\\
place blue\_2D\_rectangle, pick white\_3D\_cuboid, place white\_3D\_cuboid\\
Task planning is done successfully.
}

\newpage
\section{Experiments}~\label{app:Exp}
\subsection{Instances}~\label{app:instances}

\begin{figure*}[h!]
\captionsetup{skip=0pt}
    \centering
    \includegraphics[width=0.93\textwidth]{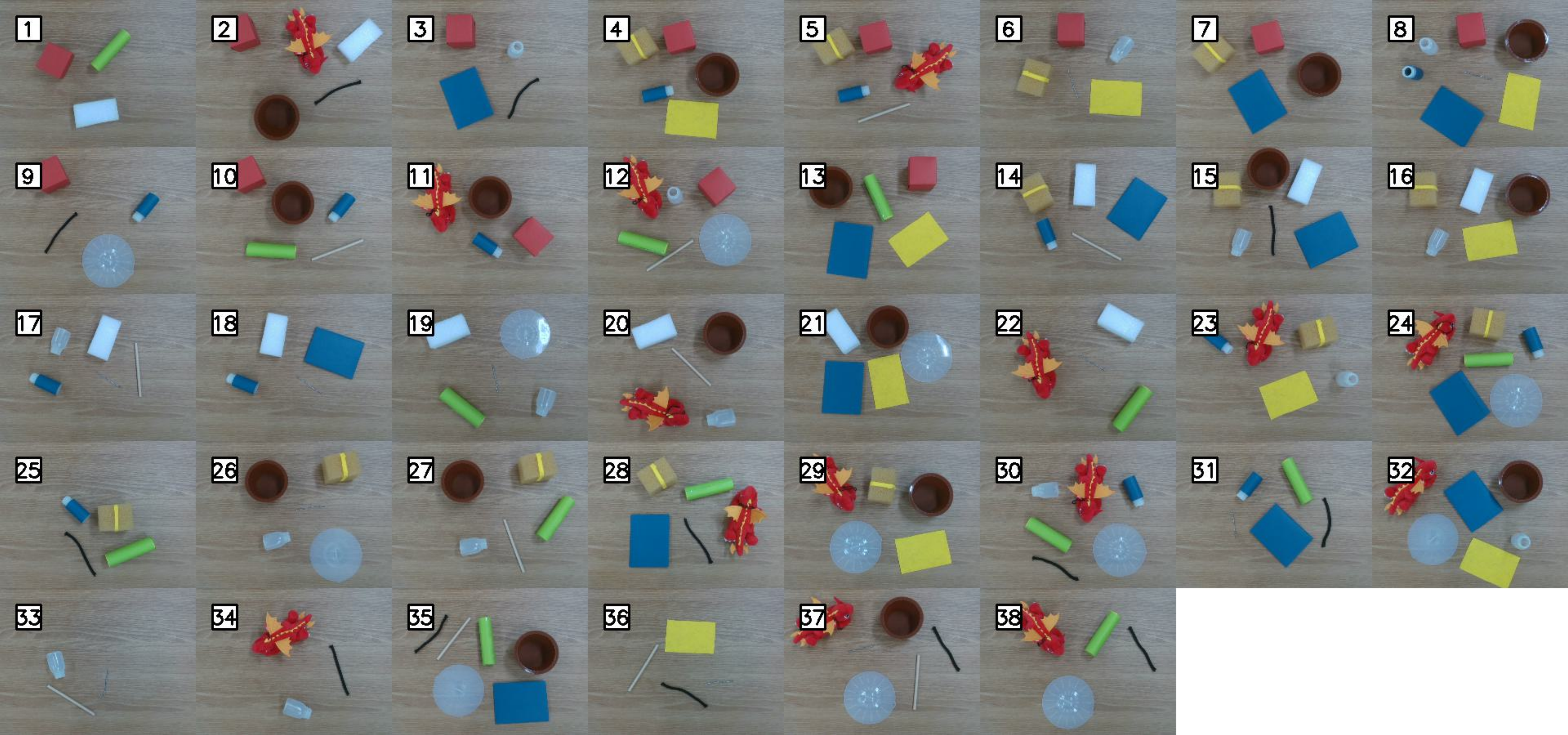}
    \caption{All 38 instances used in this experiment. All instances in this experiment were created by randomly selecting objects and refining them to ensure that no instance contains another.}    
    \label{fig:all_inst}
\end{figure*} 

Instance1: [`red \_3D \_cuboid', `white \_3D \_cuboid', `green \_3D \_cylinder'] \\
Instance2: [`red \_3D \_cuboid', `white \_3D \_cuboid', `black \_1D \_line', `red \_3D \_polyhedron', `brown \_3D \_cylinder'] \\
Instance3: [`red \_3D \_cuboid', `white \_3D \_cylinder', `blue \_2D \_rectangle', `black \_1D \_line'] \\
Instance4: [`red \_3D \_cuboid', `yellow \_3D \_cuboid', `blue \_3D \_cylinder', `yellow \_2D \_rectangle', `brown \_3D \_cylinder'] \\
Instance5: [`red \_3D \_cuboid', `yellow \_3D \_cuboid', `blue \_3D \_cylinder', `beige \_1D \_line', `red \_3D \_polyhedron'] \\
Instance6: [`red \_3D \_cuboid', `yellow \_3D \_cuboid', `white \_3D \_cylinder', `yellow \_2D \_rectangle', `gray \_1D \_line'] \\
Instance7: [`red \_3D \_cuboid', `yellow \_3D \_cuboid', `blue \_2D \_rectangle', `brown \_3D \_cylinder'] \\
Instance8: [`red \_3D \_cuboid', `blue \_3D \_cylinder', `white \_3D \_cylinder', `blue \_2D \_rectangle', `yellow \_2D \_rectangle', `gray \_1D \_line', `brown \_3D \_cylinder'] \\
Instance9: [`red \_3D \_cuboid', `blue \_3D \_cylinder', `transparent \_2D \_circle', `black \_1D \_line'] \\
Instance10: [`red \_3D \_cuboid', `blue \_3D \_cylinder', `beige \_1D \_line', `brown \_3D \_cylinder', `green \_3D \_cylinder'] \\
Instance11: [`red \_3D \_cuboid', `blue \_3D \_cylinder', `red \_3D \_polyhedron', `brown \_3D \_cylinder'] \\
Instance12: [`red \_3D \_cuboid', `white \_3D \_cylinder', `transparent \_2D \_circle', `beige \_1D \_line', `red \_3D \_polyhedron', `green \_3D \_cylinder'] \\
Instance13: [`red \_3D \_cuboid', `blue \_2D \_rectangle', `yellow \_2D \_rectangle', `brown \_3D \_cylinder', `green \_3D \_cylinder'] \\
Instance14: [`white \_3D \_cuboid', `yellow \_3D \_cuboid', `blue \_3D \_cylinder', `blue \_2D \_rectangle', `beige \_1D \_line'] \\
Instance15: [`white \_3D \_cuboid', `yellow \_3D \_cuboid', `white \_3D \_cylinder', `blue \_2D \_rectangle', `black \_1D \_line', `brown \_3D \_cylinder'] \\
Instance16: [`white \_3D \_cuboid', `yellow \_3D \_cuboid', `white \_3D \_cylinder', `yellow \_2D \_rectangle', `brown \_3D \_cylinder'] \\
Instance17: [`white \_3D \_cuboid', `blue \_3D \_cylinder', `white \_3D \_cylinder', `beige \_1D \_line', `gray \_1D \_line'] \\
Instance18: [`white \_3D \_cuboid', `blue \_3D \_cylinder', `blue \_2D \_rectangle', `gray \_1D \_line'] \\
Instance19: [`white \_3D \_cuboid', `white \_3D \_cylinder', `transparent \_2D \_circle', `gray \_1D \_line', `green \_3D \_cylinder'] \\
Instance20: [`white \_3D \_cuboid', `white \_3D \_cylinder', `beige \_1D \_line', `red \_3D \_polyhedron', `brown \_3D \_cylinder'] \\
Instance21: [`white \_3D \_cuboid', `blue \_2D \_rectangle', `yellow \_2D \_rectangle', `transparent \_2D \_circle', `brown \_3D \_cylinder'] \\
Instance22: [`white \_3D \_cuboid', `gray \_1D \_line', `red \_3D \_polyhedron', `green \_3D \_cylinder'] \\
Instance23: [`yellow \_3D \_cuboid', `blue \_3D \_cylinder', `white \_3D \_cylinder', `yellow \_2D \_rectangle', `red \_3D \_polyhedron'] \\
Instance24: [`yellow \_3D \_cuboid', `blue \_3D \_cylinder', `blue \_2D \_rectangle', `transparent \_2D \_circle', `red \_3D \_polyhedron', `green \_3D \_cylinder'] \\
Instance25: [`yellow \_3D \_cuboid', `blue \_3D \_cylinder', `black \_1D \_line', `green \_3D \_cylinder'] \\
Instance26: [`yellow \_3D \_cuboid', `white \_3D \_cylinder', `transparent \_2D \_circle', `gray \_1D \_line', `brown \_3D \_cylinder'] \\
Instance27: [`yellow \_3D \_cuboid', `white \_3D \_cylinder', `beige \_1D \_line', `brown \_3D \_cylinder', `green \_3D \_cylinder'] \\
Instance28: [`yellow \_3D \_cuboid', `blue \_2D \_rectangle', `black \_1D \_line', `red \_3D \_polyhedron', `green \_3D \_cylinder'] \\
Instance29: [`yellow \_3D \_cuboid', `yellow \_2D \_rectangle', `transparent \_2D \_circle', `red \_3D \_polyhedron', `brown \_3D \_cylinder'] \\
Instance30: [`blue \_3D \_cylinder', `white \_3D \_cylinder', `transparent \_2D \_circle', `black \_1D \_line', `red \_3D \_polyhedron', `green \_3D \_cylinder'] \\
Instance31: [`blue \_3D \_cylinder', `blue \_2D \_rectangle', `black \_1D \_line', `gray \_1D \_line', `green \_3D \_cylinder'] \\
Instance32: [`white \_3D \_cylinder', `blue \_2D \_rectangle', `yellow \_2D \_rectangle', `transparent \_2D \_circle', `red \_3D \_polyhedron', `brown \_3D \_cylinder'] \\
Instance33: [`white \_3D \_cylinder', `beige \_1D \_line', `gray \_1D \_line'] \\
Instance34: [`white \_3D \_cylinder', `black \_1D \_line', `red \_3D \_polyhedron'] \\
Instance35: [`blue \_2D \_rectangle', `transparent \_2D \_circle', `beige \_1D \_line', `black \_1D \_line', `brown \_3D \_cylinder', `green \_3D \_cylinder'] \\
Instance36: [`yellow \_2D \_rectangle', `beige \_1D \_line', `black \_1D \_line', `gray \_1D \_line'] \\
Instance37: [`transparent \_2D \_circle', `beige \_1D \_line', `black \_1D \_line', `gray \_1D \_line', `red \_3D \_polyhedron', `brown \_3D \_cylinder'] \\
Instance38: [`transparent \_2D \_circle', `black \_1D \_line', `red \_3D \_polyhedron', `green \_3D \_cylinder'] \\

\subsection{Property Probing Execution}~\label{app:property_probing}
\begin{figure}[h]
    \centering
    \includegraphics[width=0.98\textwidth]{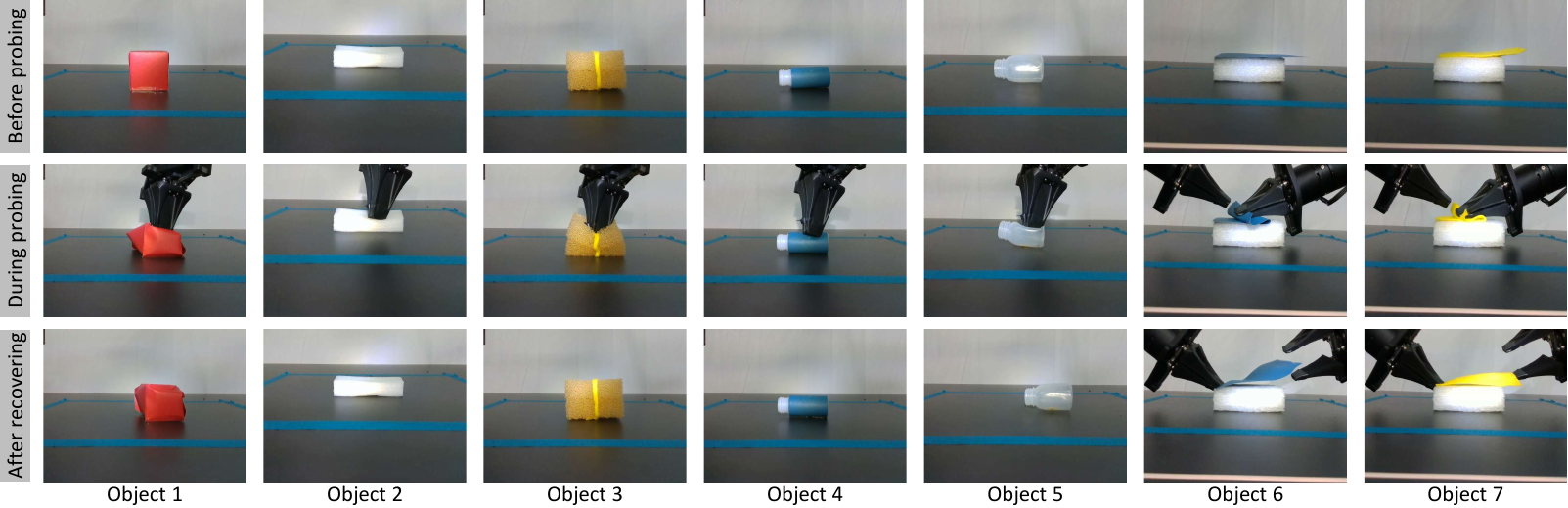}
    \hspace{0.05\linewidth}
    \includegraphics[width=0.98\textwidth]{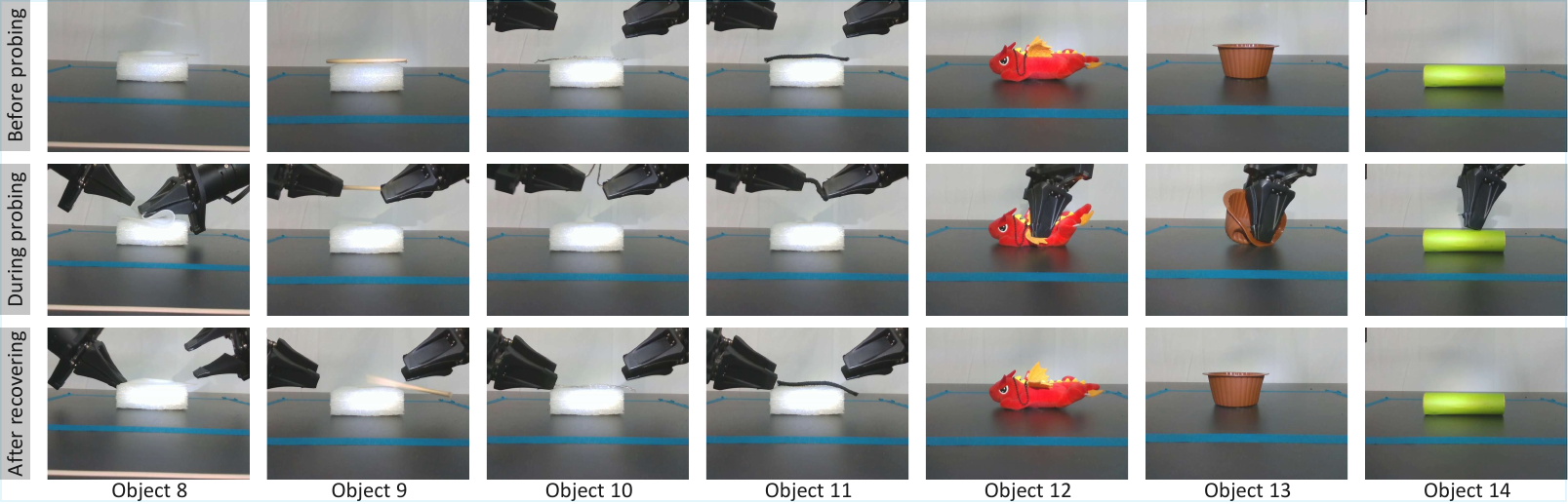}
    \caption{The implementation of the property probing. Images are obtained using a horizontal camera view without extrinsic calibration.}
    \label{fig:real_property_probing}
\end{figure}

We automate the property probing process as shown in~\ref{fig:real_property_probing}. Please refer to the supplementary video for further details.

\newpage
\subsection{Task Planning Demo}~\label{app:task_planning_demo}
\begin{figure*}[h!]
\captionsetup{skip=0pt}
    \centering
    \includegraphics[width=0.93\textwidth]{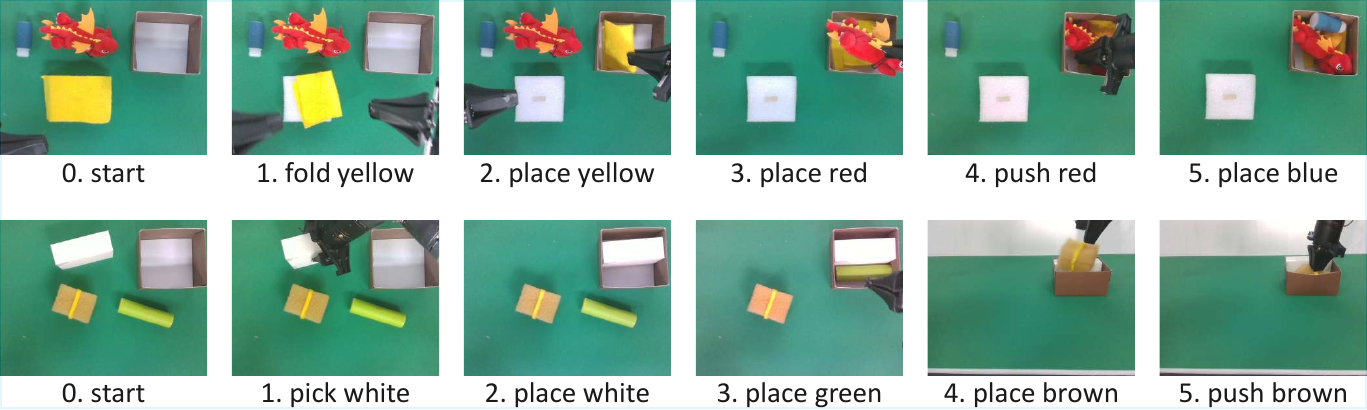}
    \caption{The real-world execution of task planning which leverages probed properties. Given the instruction ``packing all object in the box'', the first task ~~~  }    
    \label{fig:task_planning_demo}
\end{figure*}

We demonstrate that the probed properties can be utilized in bin-packing scenarios. As shown in Fig.~\ref{fig:task_planning_demo}, the robot folds the foldable object that does not initially fit into the bin. Additionally, it can push compressible objects to secure a more space in the bin.

\end{document}